\documentclass{article}

\usepackage{microtype}
\usepackage{graphicx}
\usepackage{subcaption}
\usepackage{booktabs,adjustbox,placeins} 

\usepackage{hyperref}

\usepackage[preprint]{icml2026}

\usepackage{amsmath}
\usepackage{amssymb}
\usepackage{mathtools}
\usepackage{amsthm}

\usepackage[capitalize,noabbrev]{cleveref}

\theoremstyle{plain}

\theoremstyle{definition}

\theoremstyle{remark}

\usepackage[textsize=tiny]{todonotes}

\icmltitlerunning{Written Once, Read Differently}

\begin{document}

\twocolumn[
  \icmltitle{Written Once, Read Differently: \\
  How Language Models’ Own Answers Steer Their Next Ones}



  \icmlsetsymbol{equal}{*}

  \begin{icmlauthorlist}
    \icmlauthor{Marcus Armstrong}{uh}
  \end{icmlauthorlist}

  \icmlaffiliation{uh}{Department of Computer Science, University of Houston, Houston, TX, USA}

  \icmlcorrespondingauthor{Marcus Armstrong}{miarmstr@cougarnet.uh.edu}

  \icmlkeywords{mechanistic interpretability, multi-turn consistency, linear representations, activation ablation, attention heads, attention sinks, language model reliability}

  \vskip 0.3in
]



\printAffiliationsAndNotice{}  

\begin{abstract}
When a language model answers a yes/no question, that answer changes how it
answers a different question about the same entity. We study this
\emph{self-commitment} in 22 open-weight models from four families
(0.5B--72B parameters), fixing the model's first answer and measuring the
shift in its second across 3,600 question pairs about 120 entities. Each
answer writes a \emph{commitment direction} into the residual stream. The
similarity of two questions' directions predicts their coupling on held-out
entities ($\rho = 0.43$--$0.85$), and ablating a single direction shared
across topics leaves at most 17\% of the effect in every model. How this
direction is read is far less stable: the reading circuit spreads across
more attention heads as models grow, instruction tuning makes more questions
read it in the opposite sense, and in one model changing only the system
prompt reverses the circuit's sign. Ablating the direction while sparing the
attention sink removes most of the coupling at under 1.3\% perplexity cost
in 18 of 19 instruction-tuned models, though it shifts standalone answers in
two. Without the sink spared, the direction's alignment with the leading
principal component of ordinary text predicts which models pay a high cost.
Written once, the commitment is read differently across scale, training and
prompt.
\end{abstract}

\section{Introduction}
\label{sec:intro}

Language models are now routinely deployed in settings where they make many
related judgments in sequence: an assistant answers a series of questions,
an agent evaluates the consequences of its own earlier decisions, and a
grader scores one item after another. In each case, the model's earlier outputs
sit in context when it produces later ones. Whether and how those earlier
outputs shape later judgments on \emph{different} questions is a basic
reliability question: if an early answer, right or wrong, systematically
tilts what follows, errors propagate and evaluation results become
order-dependent.

Prior work has documented that question order affects model behaviour
\citep{qqaudit2026}, and that performance degrades as information is spread
across turns \citep{laban2025lost}. A related literature on sycophancy
studies how models shift toward a user's stated views
\citep{sharma2023sycophancy}, and recent work has begun to locate the
internal mechanisms responsible \citep{persuasion2026, sycophancy2026}.
The pressure we study originates in the model's own prior output. Nothing
in the conversation asks the model to agree with itself; the earlier answer
is only present in context. Whether
this self-generated pressure shares a common mechanism across models and
scales, and whether it can be removed without cost to general ability, is
unknown.

We study \emph{self-commitment} with a minimal, controlled design. For each
of 120 entities in four domains we write six related yes/no judgments (\emph{Is chess a
sport? Is chess a game? Is chess physically demanding?}), forming directed
pairs of distinct questions. We place a first answer, \emph{Yes} or
\emph{No}, in the model's own prior turn and measure the resulting shift in
log-odds for a second, distinct question. We call this shift the
\emph{coupling}. In the primary framing the model answers as a respondent to an opinion
survey; an assistant framing and a true/false framing serve as robustness
checks. Because the two conditions differ in exactly one token,
every difference in the model's computation is attributable to that token.
We apply this design to 22 open-weight models spanning four families (Qwen,
Llama-3.1, Gemma-2, and Mistral) and 0.5B to 72B parameters, including three
matched base and instruction-tuned pairs. All layers, heads and bands used
for analysis are selected on one half of the entities and evaluated on the
other; every result is replicated over three independent entity seeds.

The results support a consistent picture: \textbf{within each model,
commitment is written into a single direction shared across topics, and
this direction is read differently depending on model size and training}.
The cosine similarity between two questions' commitment directions, each
estimated from that question alone, predicts the sign and magnitude of
their coupling on held-out entities ($\rho = 0.43$--$0.85$), and continues to predict it
after controlling for the similarity of the two questions' text and of the
model's states before answering (partial $\rho = 0.27$--$0.68$). This
prediction improves with scale (rank correlation with parameter count
$0.70$, $p = 0.001$, across the 19 instruction-tuned models). 
Mean-ablating that single shared direction across
a band of layers chosen from each model's layer profile leaves at most $17\%$ of the coupling in
every model, while ablating a random direction leaves it largely intact. The result
holds at every confidence level and in both full and 4-bit precision. The circuit that reads this direction varies.
In base models and the three smallest instruction-tuned models, the ten
highest-scoring attention heads carry all of the coupling, whereas at
70--72B parameters they carry $28$--$40\%$. Instruction tuning also changes
the reader. A fixed step along the direction pushes the second answer the
opposite way for $0$--$21\%$ of questions in the three base models, for
$14$--$49\%$ in their instruction-tuned counterparts, and for $96\%$ in
Qwen3-4B-Instruct-2507. 
Finally, ablation that spares the first token, the attention sink where
models concentrate large activations \citep{xiao2024attentionsinks,
sun2024massive}, leaves at most $14\%$ of the coupling, raises perplexity by
at most $1.3\%$ in 18 of the 19 instruction-tuned models, and preserves
standalone answers to at least $90\%$ of questions in 16 of them.

\paragraph{Contributions.}
\begin{enumerate}
    \item A controlled causal measurement of self-commitment across 22
    open-weight models spanning 0.5B--72B parameters, with exact one-token
    interventions and held-out evaluation on three independent seeds.
    \item Evidence that self-commitment depends on a single direction shared
    across topics within each model, extending the single-direction structure identified for
    refusal \citep{arditi2024refusal} to a distinct behaviour and
    replicating it across scale, model families, confidence levels, and
    numerical precision.
    \item A dissociation between a fixed write-side representation and a
    variable read-side circuit: the direction's geometry predicts coupling in every model, while the
    circuit that reads it becomes more distributed with scale and responds
    negatively for more questions after instruction tuning.
    \item A removal of self-commitment by mean-ablation of the direction that
    spares the attention sink, at a perplexity cost of at most $1.3\%$ in
    18 of 19 instruction-tuned models and MMLU changes of at most $3.5$
    points, together with its limits: it shifts standalone answers in two
    Qwen3 models, and its cost without sparing the sink is predicted by the
    direction's alignment with the leading principal component of ordinary
    text.
\end{enumerate}

\begin{figure}[t]
    \centering
    \includegraphics[width=\linewidth]{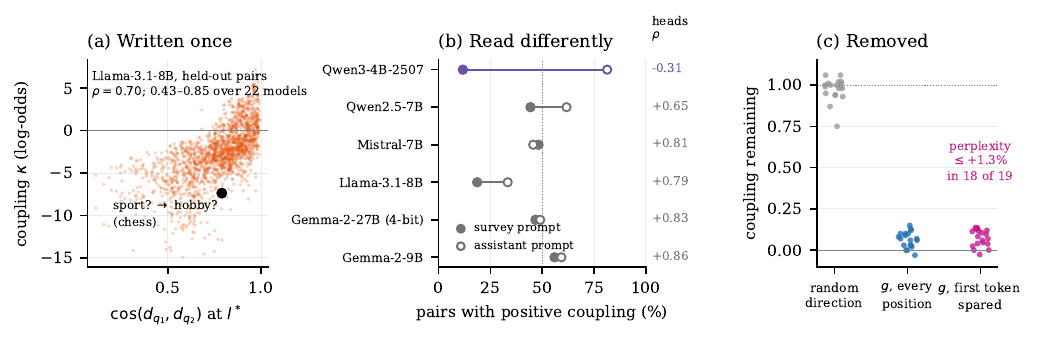}
    \caption{\textbf{Written once, read differently.} (a)~The similarity of two
        questions' commitment directions predicts their coupling on held-out pairs
        (Llama-3.1-8B; the marked pair is \emph{Is chess a sport?} followed by
        \emph{Is chess mainly a hobby?}). (b)~Pairs with positive coupling under
        two system prompts; numbers give the correlation between the two
        head-attribution maps. In Qwen3-4B-Instruct-2507 the prompt alone reverses
        the reading circuit. (c)~Coupling remaining in the 19 instruction-tuned
        models after ablating a random direction, the shared direction $g$ at every
        position, or $g$ with the first token spared.}
    \label{fig:overview}
\end{figure}

\section{Related Work}
\label{sec:related}

\paragraph{Multi-turn reliability and self-correction.}
Several lines of work document that language model outputs are sensitive to
context in ways that degrade reliability. Question order effects in yes/no
evaluations have been measured at scale \citep{qqaudit2026}, and performance drops when a task's specification is revealed over several
conversation turns instead of at once \citep{laban2025lost}. Models also
fail to self-correct reasoning errors without external feedback
\citep{huang2023cannot}, so their earlier outputs constrain later ones
even when those outputs are wrong. We ask where this constraint is
computed and whether it can be removed.

\paragraph{Sycophancy and user-driven context pressure.}
Sycophancy, the tendency of models to shift their answers toward a user's
expressed views, has been measured behaviourally \citep{sharma2023sycophancy}
and more recently located inside the model: attention heads that mediate
opinion change under user pressure have been identified
\citep{persuasion2026}, and linear representations of sycophantic pressure
have been characterised \citep{sycophancy2026}. Self-commitment differs in
its source. The earlier answer belongs to the model, and neither the user
nor the system prompt asks for consistency, so methods that detect or
suppress a user's influence do not apply to it directly.

\paragraph{Linear representations and directional ablation.}
A growing body of work finds that functionally important concepts in transformer
residual streams are approximately linear. Representation engineering
\citep{zou2023repeng} treats concepts as directions in activation space and
steers them directly. The same approach has identified a single direction
mediating refusal in instruction-tuned models \citep{arditi2024refusal},
structured linear representations of truthfulness
\citep{marks2023geometry}, and directions encoding in-context task identity
\citep{hendel2023, todd2023, liu2023invectors}. 
\citet{bigelow2025belief} model in-context learning and activation steering
as updates to a shared belief over latent concepts, with in-context examples
acting as evidence and steering shifting the prior, and predict sharp
transitions when the two are combined. In our setting the model's own
earlier answer plays the role of in-context evidence. Inference-time intervention
\citep{li2023iti} applies this structure to elicit truthful responses by patching
toward truth-associated directions at inference. We find that removing a single direction per model removes self-commitment
in all 22 models we test, which extends this pattern to a cross-question,
multi-turn setting. To our knowledge, a direction whose geometry predicts
behaviour in every model while the circuit that reads it changes with scale
and training has not been reported before.

\paragraph{Mechanistic interpretability and circuit transfer across scale.}
Mechanistic interpretability characterises model behaviour in terms of
attention circuits \citep{elhage2021mathematical}. A recurring practical
question is whether such circuits transfer across model sizes or training
regimes. Our results separate the two. The geometry of the commitment direction
predicts coupling in every model and family we test, while the share of the
coupling carried by the ten highest-scoring heads falls from all of it in
the three smallest instruction-tuned models and in base models to
$28$--$40\%$ at 70--72B parameters; ten random heads carry between $-1\%$
and $10\%$. Within one model, the head map can also reverse sign when only
the system prompt changes (Qwen3-4B-Instruct-2507). Where a behaviour is
computed can therefore change across scale and context even when the
behaviour itself is stable, which limits how far small-model circuit
findings can be assumed to generalise.

\paragraph{Attention sinks and massive activations.}
Many transformers place a large share of attention on the first token, which
\citet{xiao2024attentionsinks} term the attention sink, and a few activations
of very large magnitude appear at the first token and some delimiter tokens,
where they act as fixed biases \citep{sun2024massive}. We find that this
position governs the cost of directional ablation: sparing the first token
reduces the perplexity increase from Gemma-2-27B's $1{,}440\%$ to $10\%$,
whereas sparing all massive-activation positions reduces it only to $161\%$.

\paragraph{Language models as survey respondents.}
Language models are increasingly used to simulate survey respondents
\citep{argyle2023outofone, santurkar2023whose}, and their answers are
sensitive to question order, option labelling and wording
\citep{dominguez2024questioning, tjuatja2024response}. Our primary framing
places the model in this role; we study the mechanism of one order effect
and leave its consequences for survey simulation to future work.

\section{Setup}
\label{sec:setup}

\subsection{Questions, Framings and Coupling}

We use 120 entities in four domains (activities, foods, animals and social
groups; 30 entities each). For each domain we write six yes/no questions
that apply to every entity in it, mixing categorical and evaluative
judgments (\emph{Is chess a sport? Is chess physically demanding? Are most
politicians honest?}). Every pair of distinct questions about the same
entity is asked in both orders, giving 30 directed pairs $(q_1, q_2)$ per
entity and 3{,}600 in total (Appendix~\ref{app:dataset}).

Each conversation contains a system prompt, $q_1$ as a user turn, a forced
answer $a_1 \in \{\text{Yes}, \text{No}\}$ as the model's own turn, and
$q_2$ as a second user turn, rendered with the model's chat template. In the
primary framing the system prompt asks the model to answer as a randomly
selected adult taking a short opinion survey, with a single word, Yes or No.
Two further framings serve as robustness checks: a helpful-assistant
prompt, and the survey prompt with each question restated as a statement to
be answered True or False. Each framing has two system prompts that list
the answer words in opposite orders, and every quantity is averaged over
the two.

We read the model's answer to $q_2$ as the log-odds
\begin{equation}
    \ell = \log \textstyle\sum_{t \in T_\text{Yes}} p(t)
         - \log \textstyle\sum_{t \in T_\text{No}} p(t),
\end{equation}
where $T_\text{Yes}$ and $T_\text{No}$ are the first tokens of the answer
words with and without a leading space and in either case. The
\textbf{coupling} of a directed pair is
\begin{equation}
    \kappa(q_1, q_2) = \ell(q_2 \mid a_1 = \text{Yes})
                     - \ell(q_2 \mid a_1 = \text{No}).
    \label{eq:coupling}
\end{equation}
A positive coupling means the first answer pulls the second in the same
direction. For every pair and system prompt we verify that the two
conversations differ in exactly one token, the answer, and are identical
before it, so any difference in the model's computation originates at that
token.

\subsection{Commitment Directions}

For each question $q$ we run the conversation up to and including the
forced answer, for $a = \text{Yes}$ and $a = \text{No}$, and take the
residual-stream states at the answer token. The \textbf{commitment
direction} of $q$ at hidden layer $l$ is
\begin{equation}
    d_q(l) = h_l(\text{Yes}; q) - h_l(\text{No}; q),
    \label{eq:direction}
\end{equation}
averaged over the two system prompts, where $l = 0$ is the embedding and
$l \geq 1$ the output of block $l$. Because the conversation up to the
answer contains only $q$, $d_q$ is estimated from that question alone. Our
geometry predictor for a pair is the cosine similarity
$\cos(d_{q_1}(l^*), d_{q_2}(l^*))$, compared with the coupling by Spearman's
$\rho$. The layer $l^*$ maximises $|\rho|$ on the selection half of the
entities (\S\ref{sec:protocol}). As a control we also compute the partial
correlation after removing the similarity of the two questions' states
before answering and of their question text.

We estimate a \textbf{shared direction} $g$ in two ways. For analyses at
$l^*$, $g$ is the normalised mean of $d_q(l^*)$ over questions from the
three domains other than the pair's own (leave-one-domain-out). For
interventions across many layers, $g(l)$ is the normalised mean of $d_q(l)$
over all questions about selection-half entities, so held-out pairs never
contribute to it. Projecting $d_q$ onto $g$ splits it into a shared
component $(d_q \cdot g)\,g$ and a question-specific remainder.

\subsection{Interventions and Causal Shares}

We report every causal effect as a \textbf{share}: the least-squares slope
through the origin of the edited effect on $\kappa$ across held-out pairs,
so that 1 means the edit reproduces, or leaves, the whole coupling. A
\emph{patch} adds a vector at the first answer token at one layer of the No
conversation; patching $d_q(l)$ restores the Yes state at that token from
layer $l$ onward. An \emph{ablation} replaces the projection of the state on
a unit direction $u$ with a reference value $m$,
$h \leftarrow h - ((h \cdot u) - m)\,u$, in both conversations. The
\emph{layer profile} patches $d_q$ at one of 13 evenly spaced layers. Its
share is 1 at the embedding and falls as later positions read the
commitment from the answer token (Fig.~\ref{fig:profile}). The
\emph{readout band} runs from the last profiled layer with share at least
0.9 to the first later layer with share at most 0.1. For control
(\S\ref{sec:control}) we ablate $g(l)$ at every position and every layer of
the band, with $m$ the mean projection over 100 WikiText-2 passages
excluding the first token, and measure side effects on perplexity
(WikiText-2 test passages of 256 tokens), zero-shot MMLU accuracy, and the
answers to each question asked on its own.

\subsection{Models}

We evaluate 22 open-weight models from four families (Qwen2.5 and Qwen3,
Llama-3.1, Gemma-2 and Mistral) from 0.5B to 73B parameters: 19
instruction-tuned models and three base models paired with instruction-tuned
versions (Table~\ref{tab:models}). Qwen3 models are hybrid reasoning models,
and we disable thinking through the chat template. Qwen3-4B-Instruct-2507 is
a later post-trained release built on the same base model as Qwen3-4B.
Models up to 32B run in bfloat16 on one or two NVIDIA RTX 6000 Ada GPUs
(48\,GB each); the 70B and 72B models use 4-bit NF4 weights with bfloat16
computation. Quantisation is weight-only, so each token's computation is
independent of the batch, as the exact checks below require. Five models
are also run in 4-bit as a check (Qwen2.5-14B and 32B, Qwen3-14B and 32B,
Gemma-2-27B): held-out $\rho$ differs from bfloat16 by at most $0.09$, and
coupling remaining after band ablation by at most $0.03$
(Appendix~\ref{app:4bit}). For the 12 models up to 9B we also evaluate the
same architecture with random weights.

\begin{table}[t]
\centering
\caption{Models evaluated. $\dagger$ marks base models; all others are
instruction-tuned. $^\ast$ marks models also run with 4-bit weights as a
quantisation check. Qwen3-4B-Instruct-2507 is a later post-trained release
of the same base model as Qwen3-4B and is discussed in \S\ref{sec:reader}.}
\label{tab:models}
\small
\begin{tabular}{llcl}
\toprule
Family & Model & Params & Precision \\
\midrule
Qwen2.5 & Qwen2.5-0.5B-Instruct         & 0.5B  & bf16  \\
        & Qwen2.5-1.5B-Instruct         & 1.5B  & bf16  \\
        & Qwen2.5-3B-Instruct           & 3.1B  & bf16  \\
        & Qwen2.5-7B-Instruct           & 7.6B  & bf16  \\
        & Qwen2.5-7B$^\dagger$          & 7.6B  & bf16  \\
        & Qwen2.5-14B-Instruct$^\ast$   & 14.8B & bf16  \\
        & Qwen2.5-32B-Instruct$^\ast$   & 32.8B & bf16  \\
        & Qwen2.5-72B-Instruct          & 72.7B & 4-bit \\
\midrule
Qwen3   & Qwen3-0.6B                    & 0.6B  & bf16  \\
        & Qwen3-1.7B                    & 1.7B  & bf16  \\
        & Qwen3-4B                      & 4.0B  & bf16  \\
        & Qwen3-4B-Base$^\dagger$       & 4.0B  & bf16  \\
        & Qwen3-4B-Instruct-2507        & 4.0B  & bf16  \\
        & Qwen3-8B                      & 8.2B  & bf16  \\
        & Qwen3-14B$^\ast$              & 14.8B & bf16  \\
        & Qwen3-32B$^\ast$              & 32.8B & bf16  \\
\midrule
Llama   & Llama-3.1-8B-Instruct         & 8.0B  & bf16  \\
        & Llama-3.1-70B-Instruct        & 70.6B & 4-bit \\
\midrule
Gemma-2 & Gemma-2-9B-it                 & 9.2B  & bf16  \\
        & Gemma-2-27B-it$^\ast$         & 27.2B & bf16  \\
\midrule
Mistral & Mistral-7B-Instruct-v0.3      & 7.2B  & bf16  \\
        & Mistral-7B-v0.3$^\dagger$     & 7.2B  & bf16  \\
\bottomrule
\end{tabular}
\end{table}

\subsection{Evaluation Protocol}
\label{sec:protocol}

Entities are divided into three disjoint folds of 40 (10 per domain). Each
seed uses one fold, so the three seeds are independent replications on
different entities, with 1{,}200 directed pairs each. Within a seed,
entities are split at random into a selection half and a test half. The
layer $l^*$, the layer profile and readout band, the attention heads, and
the band direction $g$ are chosen on the selection half, and every reported
number is computed on the 600 test pairs or a fixed subset of them
(Appendix~\ref{app:samples}). The leave-one-domain-out $g$ uses directions,
never couplings, of test-half questions from other domains. We report means
$\pm$ s.d.\ over the three seeds; 95\% intervals from a cluster bootstrap
over entities (1{,}000 resamples) are given in the appendix. Correlations
across models use Spearman's $\rho$ with permutation $p$-values (10{,}000
permutations).

\subsection{Sanity Checks}

Two checks confirm that the interventions are exact. Patching $d_q(0)$ into
the No conversation turns the answer token's embedding into that of the Yes
conversation and reproduces the coupling (share $0.996$--$1.011$). Ablating
$d_q$ at the answer token at every layer removes it (share remaining
$-0.003$ to $0.004$). Both are within $0.05$ of their targets for every
model, framing and seed (Appendix~\ref{app:checks}).

\begin{figure}[t]
    \centering
    \includegraphics[width=\linewidth]{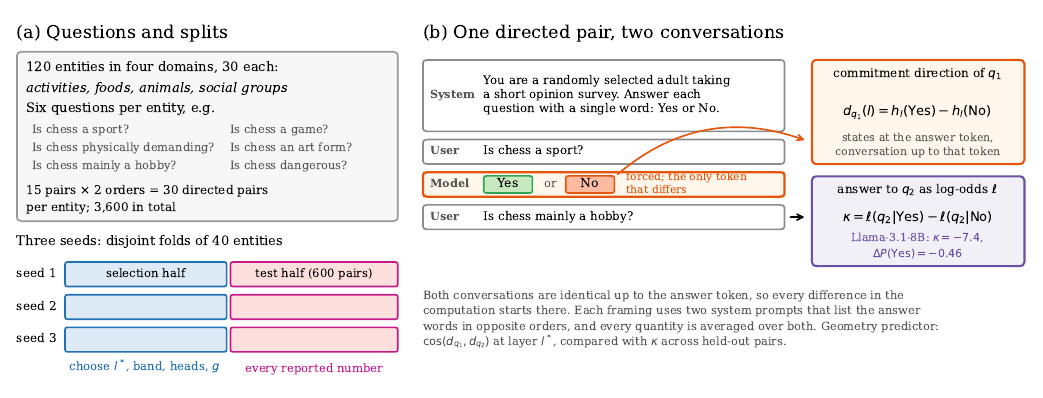}
    \caption{\textbf{Experimental design.} For each directed pair $(q_1, q_2)$
        about one entity, the model's own first answer is fixed to Yes or No, and
        the coupling $\kappa$ (Eq.~\ref{eq:coupling}) is the difference in the
        log-odds of its answer to $q_2$. The two conversations differ only in the
        answer token. The commitment direction $d_q$ (Eq.~\ref{eq:direction}) is the
        difference between the residual-stream states at that token.}
    \label{fig:design}
\end{figure}

\section{Results}
\label{sec:results}

\subsection{Geometry Predicts the Coupling}
\label{sec:geometry}

For each held-out pair we compare the cosine similarity of the two
questions' commitment directions at layer $l^*$ with the pair's coupling.
Across the 22 models the correlation is $\rho = 0.43$--$0.85$
(Table~\ref{tab:main}). It improves with scale across the 19
instruction-tuned models (rank correlation with parameter count $0.70$,
$p = 0.001$). Within Qwen3 it rises from $0.43$ at 0.6B to $0.83$ at 32B
with one decrease at 8B, and within Llama-3.1 from $0.70$ at 8B to $0.85$
at 70B. Base models give lower values ($0.45$--$0.61$) than their
instruction-tuned counterparts ($0.72$--$0.81$). The correlation remains
after controlling for the similarity of the two questions' text and of the
model's states before answering (partial $\rho = 0.27$--$0.68$), so it is
not explained by questions that are merely similar. Models with random
weights give $|\rho| \le 0.13$ in 10 of 12 cases; the two exceptions,
Qwen2.5-3B ($-0.31$) and Qwen3-4B-Instruct-2507 ($0.19$), are smaller in
magnitude than every trained model.

\begin{figure}[t]
    \centering
    \includegraphics[width=\linewidth]{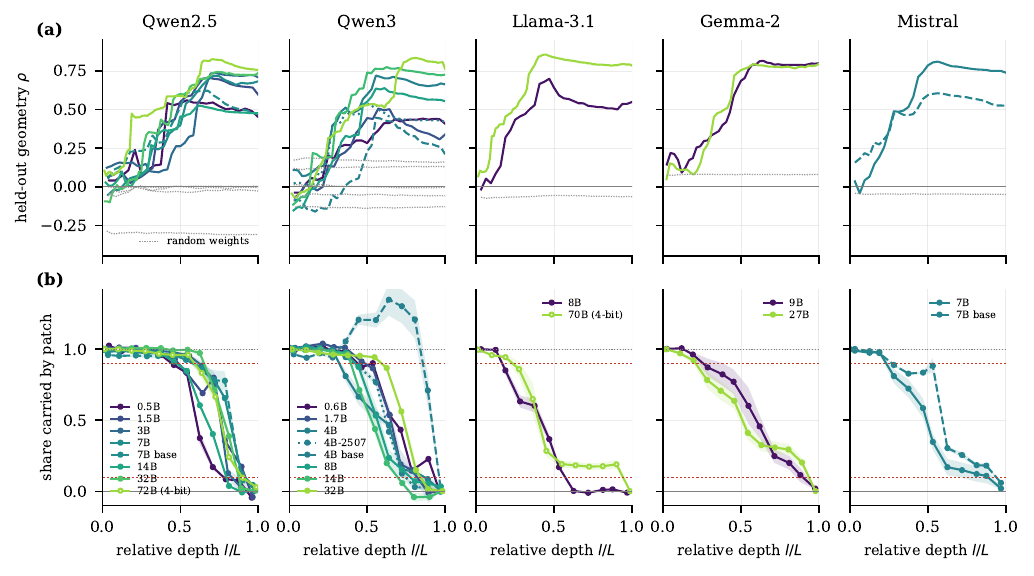}
    \caption{\textbf{Geometry and location of the commitment by layer.}
        (a)~Held-out Spearman correlation between direction similarity and
        coupling at each layer; dotted grey lines are the same architectures with
        random weights. (b)~Share of the coupling reproduced by patching $d$ at the
        first answer token at a single layer (selection-half entities; shading
        $\pm 1$ s.d.\ over seeds). Red dashed lines mark the 0.9 and 0.1 thresholds
        that define the readout band. Dashed: base models; dotted: Qwen3-4B-2507;
        hollow markers: 4-bit weights. All curves are means over three seeds.}
    \label{fig:profile}
\end{figure}

Figure~\ref{fig:profile}a shows the correlation at every layer. It is near
zero at the embedding, rises through the first half of the network and
stays high to the last layer; the selected layer $l^*$ lies inside the
readout band in 62 of 66 model--seed runs. Figure~\ref{fig:profile}b shows
where the commitment leaves the answer token. In the Qwen models the share
carried by a patch stays near 1 until 35--65\% of depth and then falls
steeply. In Llama-3.1, Gemma-2 and Mistral it begins to fall at 15--30\% of
depth and declines gradually, which gives these models wide readout bands.
In Qwen3-4B-Base the share exceeds 1 at intermediate depth (up to $1.35$),
and in Llama-3.1-70B about $0.18$ of the coupling is still read from the
answer token in the final layers.

\begin{table*}[t]
\centering
\caption{Per-model results on held-out pairs (primary framing; means over three seeds, s.d.\ in Appendix~\ref{app:fullresults}). $|\Delta P|$: mean absolute change in the probability of Yes. Pos.: pairs with positive coupling (\%). $\rho$: Spearman correlation between direction similarity and coupling; partial: after controlling for question-text and pre-answer similarity. Patch: share carried by patching $d$ at $l^*$. Heads: share carried by the ten highest-scoring heads. $\kappa_g$, $\kappa_\text{rand}$: share remaining after mean-ablating $g$ or a random direction at every position of the readout band. Read neg.: questions for which a fixed step along $g$ moves the answer the opposite way (\%). $\dagger$ base model. $\S$ 4-bit weights. $\ddagger$ from the 200-pair sink-test run; base models have no random-direction control. Dashes: not measured.}
\label{tab:main}
\small
\setlength{\tabcolsep}{4pt}
\begin{tabular}{llcccccccccc}
\toprule
Family & Model & Params (B) & $|\Delta P|$ & Pos. & $\rho$ & Partial $\rho$ & Patch & Heads & $\kappa_g$ & $\kappa_\text{rand}$ & Read neg. \\
\midrule
Qwen2.5  & 0.5B-Instruct & 0.5 & 0.14 & 71 & 0.56 & 0.50 & 0.83 & 1.07 & 0.00 & 0.75 & 28 \\
         & 1.5B-Instruct & 1.5 & 0.22 & 18 & 0.70 & 0.49 & 0.34 & 1.44 & 0.08 & 0.87 & 79 \\
         & 3B-Instruct & 3.1 & 0.13 & 33 & 0.73 & 0.36 & 0.55 & 0.70 & 0.10 & 0.95 & 71 \\
         & 7B-Instruct & 7.6 & 0.09 & 44 & 0.72 & 0.45 & 0.73 & 0.41 & 0.13 & 0.99 & 43 \\
         & 7B$^\dagger$ & 7.6 & 0.09 & 63 & 0.61 & 0.52 & 0.91 & 1.06 & 0.15$^\ddagger$ & -- & 21 \\
         & 14B-Instruct & 14.8 & 0.12 & 16 & 0.53 & 0.27 & 0.62 & 0.61 & 0.06 & 0.99 & 89 \\
         & 32B-Instruct & 32.8 & 0.07 & 20 & 0.74 & 0.54 & 0.69 & 0.58 & 0.15 & 0.98 & 94 \\
         & 72B-Instruct$^\S$ & 72.7 & 0.12 & 36 & 0.81 & 0.66 & 0.72 & 0.28 & 0.10 & 1.00 & -- \\
\midrule
Qwen3    & 0.6B & 0.6 & 0.71 & 100 & 0.43 & 0.27 & 0.39 & 1.06 & 0.06 & 1.02 & 0 \\
         & 1.7B & 1.7 & 0.09 & 44 & 0.50 & 0.45 & 0.53 & 0.61 & 0.07 & 0.94 & 41 \\
         & 4B & 4.0 & 0.17 & 52 & 0.72 & 0.64 & 0.50 & 0.69 & 0.03 & 0.93 & 14 \\
         & 4B$^\dagger$ & 4.0 & 0.07 & 74 & 0.45 & 0.51 & 1.21 & 1.34 & $-$0.15$^\ddagger$ & -- & 0 \\
         & 4B-Instruct-2507 & 4.0 & 0.17 & 12 & 0.54 & 0.54 & 0.87 & 1.24 & 0.03 & 1.06 & 96 \\
         & 8B & 8.2 & 0.09 & 45 & 0.64 & 0.55 & 0.49 & 0.48 & 0.12 & 1.00 & 25 \\
         & 14B & 14.8 & 0.08 & 41 & 0.77 & 0.68 & 0.32 & 0.59 & 0.09 & 1.02 & 59 \\
         & 32B & 32.8 & 0.09 & 47 & 0.83 & 0.61 & 0.13 & 0.41 & 0.07 & 1.06 & 67 \\
\midrule
Llama-3.1 & 8B-Instruct & 8.0 & 0.17 & 19 & 0.70 & 0.56 & 0.41 & 0.79 & 0.02 & 0.98 & 88 \\
         & 70B-Instruct$^\S$ & 70.6 & 0.06 & 36 & 0.85 & 0.63 & 0.41 & 0.40 & 0.10 & 1.01 & -- \\
\midrule
Gemma-2  & 9B-it & 9.2 & 0.06 & 56 & 0.80 & 0.55 & 0.38 & 0.42 & 0.03 & 1.01 & 33 \\
         & 27B-it & 27.2 & 0.07 & 47 & 0.79 & 0.38 & 0.25 & 0.48 & 0.00 & 0.94 & 42 \\
\midrule
Mistral  & 7B-Instruct-v0.3 & 7.2 & 0.09 & 48 & 0.81 & 0.43 & 0.35 & 0.52 & $-$0.03 & 1.00 & 49 \\
         & 7B-v0.3$^\dagger$ & 7.2 & 0.15 & 93 & 0.61 & 0.46 & 1.01 & 1.24 & 0.10$^\ddagger$ & -- & 0 \\
\bottomrule
\end{tabular}
\end{table*}

\subsection{Removing One Direction Removes the Coupling}
\label{sec:direction}

We mean-ablate $g(l)$ at every position and every layer of the readout
band and measure the share of the coupling that remains on held-out pairs;
as a specificity control we ablate a random unit direction in the same
layers. Ablating $g$ leaves between $-0.03$ and $0.15$ of the coupling in
the 19 instruction-tuned models, whereas the random direction leaves
$0.75$--$1.06$ (Table~\ref{tab:main}). In the three base models, measured
on 200 pairs, ablating $g$ leaves $-0.15$ to $0.15$. The five models run in
both precisions agree to within $0.03$. Because the ablation acts at every
position, including the tokens of the second question, it removes the
direction both where the commitment is written and where it is read.
Ablating $g$ only at the answer token at $l^*$ leaves $0.16$--$0.88$ of the
coupling, consistent with Figure~\ref{fig:profile}b: by $l^*$, later
positions have already read part of the commitment.

The shared direction also accounts for nearly all of what a single edit at
$l^*$ can carry. Patching $d_q$ at $l^*$ carries $0.13$--$0.87$ of the
coupling in instruction-tuned models. When the patch is split into its
component along the leave-one-domain-out $g$ and the remainder, the
question-specific remainder carries between $-0.03$ and $0.16$.

Table~\ref{tab:confbins} stratifies held-out pairs by the model's
standalone confidence in the second answer. Band ablation of $g$ leaves
$0.06$ of the coupling on average in every bin, and at most $0.17$ for any
model in any bin, while the random direction leaves $0.99$--$1.00$, so the
removal is not an artefact of saturated answers. The geometric prediction is
weaker away from saturation: mean held-out $\rho$ is $0.53$ for $p < 0.9$
and $0.57$ for $0.9 \le p < 0.99$, against $0.75$ above $0.99$.

\begin{table}[t]
\centering
\caption{Held-out pairs by the model's standalone probability $p$ of its
answer to the second question, averaged over the 19 instruction-tuned
models (bins with fewer than 10 pairs in a model are omitted). $\kappa_g$
and $\kappa_\text{rand}$: share remaining after band ablation of $g$ or a
random direction. Per-model values are in Appendix~\ref{app:confbins}.}
\label{tab:confbins}
\small
\setlength{\tabcolsep}{4pt}
\begin{tabular}{lccccc}
\toprule
Bin & Models & Pairs & $\rho$ & $\kappa_g$ & $\kappa_\text{rand}$ \\
\midrule
All pairs           & 19 & 34{,}200 & 0.69 & 0.06 & 1.00 \\
$p < 0.90$          & 19 &  7{,}010 & 0.53 & 0.06 & 1.00 \\
$0.90 \le p < 0.99$ & 18 &  4{,}140 & 0.57 & 0.06 & 0.99 \\
$p \ge 0.99$        & 17 & 23{,}050 & 0.75 & 0.06 & 0.99 \\
\bottomrule
\end{tabular}
\end{table}

\subsection{The Reading Circuit Distributes with Scale}
\label{sec:heads}

We rank attention heads on the selection half by attribution patching: for
each head, the gradient of the answer log-odds in the No conversation times
the difference between the head's outputs in the Yes and No conversations,
summed over positions after the first answer and regressed on the pair's
coupling. On held-out pairs we then patch the outputs of the $k$
highest-ranked heads from the Yes into the No conversation at every
position after the answer token.

The ten highest-ranked heads carry all of the coupling or more in the
three smallest instruction-tuned models ($1.06$--$1.44$), in
Qwen3-4B-Instruct-2507 ($1.24$) and in the three base models
($1.06$--$1.34$). In these models, knocking out the same heads leaves a
negative coupling ($-0.36$ to $-0.15$), so the remaining heads push in the
opposite direction. In larger models the share falls to $0.41$--$0.79$ at
7--9B and $0.28$--$0.40$ at 70--72B (Fig.~\ref{fig:scaling}c), while ten
random heads carry between $-0.01$ and $0.10$ in every model. A patch of
$d_q$ at $l^*$ shows the same trend: it carries $0.91$--$1.21$ in base
models but $0.13$--$0.72$ in instruction-tuned models of 14B and above. As
models grow, the commitment is read by more heads across more layers,
while its geometry continues to predict the coupling
(Fig.~\ref{fig:scaling}a).

\subsection{Training Shapes the Reader}
\label{sec:reader}

To measure how a question reads the direction, we add a step of fixed size
along $g$ at $l^*$ (the median projection of $d_q$ on $g$) to the No
conversation and record the change in the answer to $q_2$. Averaged over
the first questions it is paired with, this is the question's \emph{read
gain}. Within each model the read gain of a pair predicts its coupling
($\rho = 0.40$--$0.95$ in 19 of the 20 models measured; $0.12$ in
Qwen3-32B), whereas the strength with which $q_1$ writes along $g$ does not
($\rho = -0.14$ to $0.29$). The read gain of $q_2$ estimated from its other
first questions also predicts the coupling of a held-out pair
($\rho = 0.17$--$0.71$). The sign of the coupling is therefore set mainly by
how the second question reads the direction.

\begin{figure}[t]
    \centering
    \includegraphics[width=\linewidth]{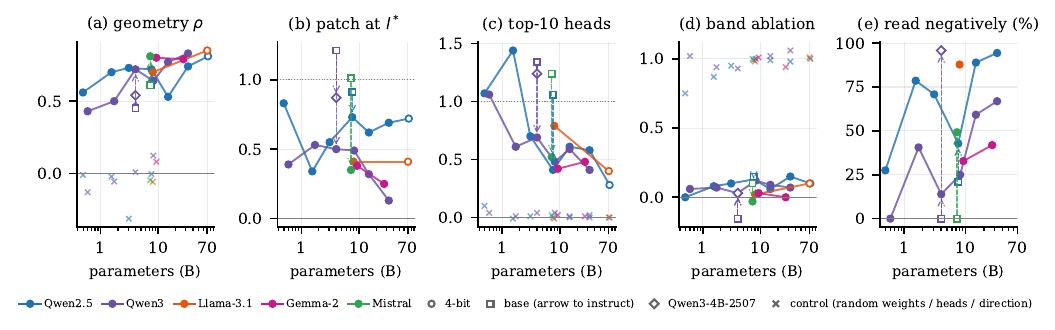}
    \caption{\textbf{Scale and training change the reader.} Per-model values
        against parameter count. Lines join the instruction-tuned models of one
        family; squares are base models with arrows to their instruction-tuned
        versions; the diamond is Qwen3-4B-Instruct-2507; hollow circles are 4-bit
        weights; crosses are controls (random weights in a, ten random heads in c,
        a random direction in d). (a)~Held-out geometry $\rho$. (b)~Share carried
        by patching $d$ at $l^*$. (c)~Share carried by the ten highest-ranked
        heads. (d)~Share remaining after band ablation of $g$. (e)~Questions read
        negatively. The 70B and 72B models have no read-gain measurement.}
    \label{fig:scaling}
\end{figure}

Instruction tuning changes this reader. The share of questions with
negative read gain is $0\%$ in base Mistral-7B and Qwen3-4B and $21\%$ in
base Qwen2.5-7B. In their instruction-tuned versions it is $49\%$, $14\%$
and $43\%$, and in Qwen3-4B-Instruct-2507, built on the same base model as
Qwen3-4B, it is $96\%$ (Fig.~\ref{fig:scaling}e). The share varies across
seeds in some models (Appendix~\ref{app:fullresults}), but it increases
after instruction tuning in all three matched pairs.

The reader can also change with the system prompt alone. For six models we
repeat the head attribution under the assistant framing and compare it with
the survey framing. In five of them the head maps correlate at
$\rho = 0.65$--$0.86$, and the survey framing's ten highest-ranked heads
still carry $0.52$--$0.88$ of the coupling. In Qwen3-4B-Instruct-2507 the
maps correlate at $\rho = -0.31$, the same ten heads carry $-0.70$ of the
coupling, and the share of pairs with positive coupling rises from $12\%$
under the survey prompt to $81\%$ under the assistant prompt, although the
direction's geometry still predicts the coupling ($\rho = 0.60$). The same
model, given a different persona, reads its earlier answer with the
opposite sign.

\section{Controlling Self-Commitment}
\label{sec:control}

Section~\ref{sec:direction} shows that ablating $g$ across the readout band
removes the coupling. Here we ask what the intervention costs when applied
at inference time, at every position of every forward pass. To keep
conditions comparable, all results in this section use the same 200
held-out pairs per seed, 100 WikiText-2 passages and 200 MMLU questions;
Table~\ref{tab:main} reports the larger control-stage sample (300 pairs,
200 passages, 500 questions).

\paragraph{Ablation at every position removes the coupling at variable cost.}
Band ablation of $g$ at every position leaves between $-0.03$ and $0.14$ of
the coupling in all 19 instruction-tuned models. Perplexity rises by at most
$9\%$ in 14 of them and by $15\%$ in Qwen3-0.6B, but by $44\%$ in Qwen3-4B,
$81\%$ in Qwen3-32B and $86\%$ in Qwen3-1.7B, and by a factor of $15.4$ in
Gemma-2-27B, which also loses $23$ points of MMLU accuracy. The intervention
also changes answers to single questions: the share of standalone answers
that keep their sign falls to $30\%$ in Qwen3-0.6B, $59\%$ in Qwen2.5-0.5B,
$66\%$ in Qwen3-4B, $72\%$ in Qwen3-1.7B and $83\%$ in Gemma-2-27B, against
$91$--$98\%$ in the other 14 models.

\paragraph{The perplexity cost comes from the first token.}
Many models direct most attention to the first token, the attention sink
\citep{xiao2024attentionsinks}, and carry a few activations of very large
magnitude there \citep{sun2024massive}. Repeating the ablation with the
first token left unchanged (Table~\ref{tab:control}) reduces the perplexity
increase to at most $1.3\%$ in 18 of the 19 models and to $10\%$ in
Gemma-2-27B, while the coupling remaining stays between $-0.03$ and $0.14$
and MMLU accuracy changes by between $-3.5$ and $+1.3$ points. Sparing the
first token also lowers the cost in the models that were cheap to begin
with, by up to $8$ percentage points (Qwen2.5-14B). Sparing every position
whose norm exceeds ten times the median is less effective, leaving
Gemma-2-27B at $2.61$ and Qwen3-1.7B at $1.21$ times baseline perplexity,
so a threshold on activation magnitude does not isolate the damaging
position.

\begin{figure}[t]
    \centering
    \includegraphics[width=\linewidth]{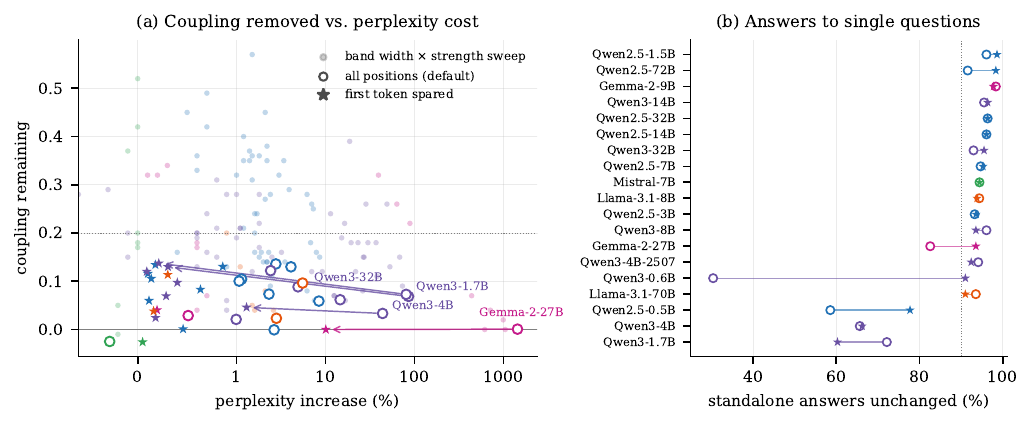}
    \caption{\textbf{Cost of removing the coupling.} (a)~Share of the coupling
        remaining against perplexity increase for the 19 instruction-tuned models.
        Small points: band width $\times$ strength sweep (17 models up to 32B,
        ablation at every position). Open circles: default band ablation at every
        position. Stars: the same ablation with the first token spared; arrows mark
        the four models with the largest cost. (b)~Share of single questions whose
        standalone answer keeps its sign, for the default ablation (circles) and
        with the first token spared (stars); the dotted line marks $90\%$. Means
        over three seeds.}
    \label{fig:tradeoff}
\end{figure}

\paragraph{Standalone answers are not fully preserved.}
Sparing the first token restores standalone answers in Qwen3-0.6B ($91\%$
unchanged) and Gemma-2-27B ($94\%$), and 16 of the 19 models keep at least
$90\%$ of them (Fig.~\ref{fig:tradeoff}b). The other three do not:
Qwen2.5-0.5B keeps $78\%$, Qwen3-4B $66\%$ and Qwen3-1.7B $60\%$. In the two
Qwen3 models the mean change is $10.8$ and $8.6$ log-odds, close to their
median standalone confidence ($13.0$ and $8.8$). Even where the sign is
kept, standalone log-odds change by $0.2$ to $5.6$ on average. The direction
$g$ therefore carries part of the model's general disposition to answer yes
or no, and removing it is not free in every model.

\begin{table*}[t]
\centering
\caption{Band ablation of $g$ at every position, and with the first token
(the attention sink) spared. PPL: perplexity ratio to baseline.
$\kappa_g$: share of coupling remaining. MMLU: change in accuracy
(points). Ans.: standalone answers that keep their sign (\%). The five
models shown have the largest cost at every position; the last row gives
the range over the other 14 instruction-tuned models. 200 held-out pairs,
100 passages and 200 MMLU questions per seed; means over three seeds.}
\label{tab:control}
\small
\begin{tabular}{lcccccccc}
\toprule
 & \multicolumn{4}{c}{All positions} & \multicolumn{4}{c}{First token spared} \\
\cmidrule(lr){2-5}\cmidrule(lr){6-9}
Model & PPL & $\kappa_g$ & MMLU & Ans. & PPL & $\kappa_g$ & MMLU & Ans. \\
\midrule
Gemma-2-27B   & 15.40 & 0.00 & $-23.2$ & 83 & 1.101 & 0.00 & $-1.7$ & 94 \\
Qwen3-1.7B    & 1.86  & 0.07 & $-1.3$  & 72 & 1.003 & 0.13 & $+0.7$ & 60 \\
Qwen3-32B     & 1.81  & 0.07 & $-0.3$  & 93 & 1.002 & 0.14 & $-0.3$ & 95 \\
Qwen3-4B      & 1.44  & 0.03 & $-1.5$  & 66 & 1.013 & 0.05 & $+1.3$ & 66 \\
Qwen3-0.6B    & 1.15  & 0.06 & $-0.8$  & 30 & 1.003 & 0.07 & $+1.3$ & 91 \\
\midrule
Other 14      & 1.00--1.08 & $-0.03$--$0.14$ & $-3.7$--$+1.0$ & 59--98
              & 1.00--1.01 & $-0.03$--$0.13$ & $-3.5$--$+1.2$ & 78--99 \\
\bottomrule
\end{tabular}
\end{table*}

\paragraph{Which models are sensitive.}
We relate $g$ to the activations of ordinary text (100 WikiText-2
passages, first token excluded). \emph{Enrichment} is the variance along
$g$ relative to the average variance per direction (about 1 for a random
direction). Gemma-2-27B has the highest enrichment of any model
($62.9 \pm 0.4$, against $0.9 \pm 0.1$ for a random direction; $1.8$--$22.8$
in the other instruction-tuned models). Across the 19 instruction-tuned
models, the squared cosine between $g$ and the first principal component of
ordinary-text activations best predicts the perplexity cost of ablation at
every position (Spearman $\rho = 0.73$, permutation $p = 0.0006$), followed
by enrichment ($\rho = 0.56$, $p = 0.013$). How far the ablation moves the
first token along $g$ does not predict the cost ($\rho = 0.01$), nor does
the first token's norm ($\rho = 0.06$), and no measure predicts the MMLU
change ($|\rho| \le 0.39$, $p \ge 0.097$). Ablation is costly in the models
where $g$ is most aligned with the dominant directions of ordinary text, and
sparing the first token removes most of that cost; Gemma-2-27B's remaining
$10\%$ is consistent with its much higher enrichment.

\paragraph{Narrower or weaker ablation does not remove the cost.}
We also sweep the width of the readout band (profile thresholds 0.9--0.1,
0.75--0.25 and 0.6--0.4) and the ablation strength
($\alpha \in \{0.5, 0.75, 1\}$) for the 17 instruction-tuned models up to
32B, with ablation at every position. For each seed we choose, on the other
two seeds, the setting with the lowest perplexity cost among those leaving
at most $20\%$ of the coupling, and report it on the held-out seed. The
chosen settings cost at most $7\%$ perplexity in 13 models, but $17\%$ in
Qwen3-4B, $55\%$ in Qwen3-32B, $61\%$ in Qwen3-1.7B and a factor of $3.9$ in
Gemma-2-27B (Fig.~\ref{fig:tradeoff}a; Appendix~\ref{app:tradeoff}). A
narrower band or a weaker ablation trades coupling against perplexity along
the same curve; sparing the first token moves these models off it.

\section{Conclusion}
\label{sec:conclusion}

A model's own earlier answer changes its answer to a different question
about the same entity, and the same structure underlies this effect in all
22 models we study. Each answer writes a commitment direction whose geometry
predicts the coupling on held-out entities ($\rho = 0.43$--$0.85$), and
ablating one shared direction across a band of layers leaves at most $17\%$
of it. How the direction is read varies: more heads read it as models grow,
more questions read it negatively after instruction tuning, and in one model
a change of system prompt reverses the sign of the reading circuit. Sparing
the attention sink makes the ablation cheap in perplexity and MMLU accuracy
for nearly all models, although it shifts standalone answers in two Qwen3
models. A stable representation therefore does not imply a stable circuit,
and circuit-level findings should be rechecked across scale, post-training
and prompts.

\paragraph{Limitations.}
First answers are inserted rather than generated, and all questions are
binary; generated or graded answers may write the commitment differently.
The questions are short judgments about 120 entities in a simulated survey,
not conversations from deployment, and what self-commitment implies for
survey simulation is left open. The base-model comparison rests on three
pairs, and the 70B and 72B models run only with 4-bit weights and without
the read-gain analysis. Read gain is correlational evidence for the role of
the reader. Finally, we measure the ablation's cost on perplexity, MMLU and
standalone answers, but not on tasks that require consistency across turns,
where removing the coupling may be undesirable.

\section*{Impact Statement}
This work studies how a language model's earlier answers shape its later
ones, a factor in the reliability of evaluations, multi-turn assistants and
the use of models as simulated survey respondents. Our results suggest that
order effects in such uses can come from model internals that differ across
models, post-training releases and system prompts, which argues for testing
them explicitly before relying on simulated responses. Some stimulus
questions ask for evaluations of occupational and social groups (for
example, whether most politicians are honest); we analyse the effect of an
earlier answer on later ones and draw no conclusions from the answers
themselves. The ablation we describe is an analysis tool. It reduces the
coupling between answers but also shifts standalone answers in some models,
and we do not recommend deploying it without task-specific evaluation.

\section*{Use of AI Assistants}
We used an AI assistant (Claude, Anthropic) during this project. It helped
write parts of the experiment code, revised text in all sections, and generated the figures.
The authors designed
the study, ran all experiments, and checked every reported number, figure
and table against the result files produced by the experiment code. The
authors take full responsibility for the content of the paper.

\bibliography{example_paper}
\bibliographystyle{icml2026}

\newpage
\appendix
\onecolumn
 
\section{Questions, Framings and Pairs}
\label{app:dataset}
 
The 120 entities fall into four domains of 30. Each domain has six yes/no question templates that apply to every entity in it (Table~\ref{tab:items}). For each entity, all 15 pairs of distinct questions are asked in both orders, giving 30 directed pairs per entity and 3{,}600 in total. The entities are shuffled once with a fixed seed and assigned to three disjoint folds of 40 (10 per domain); seed $s$ uses fold $s$, so the three seeds share no entities. Within a seed, entities are split at random into a selection half and a test half.
 
\begin{table}[!htbp]
\centering
\caption{Entities and question templates ($x$ is the entity).}
\label{tab:items}
\scriptsize
\begin{adjustbox}{max width=\linewidth}
\begin{tabular}{p{0.09\linewidth}p{0.52\linewidth}p{0.32\linewidth}}
\toprule
Domain & Entities & Questions \\
\midrule
Activities & chess, darts, poker, golf, cheerleading, esports, fishing, ballet, yoga, bowling, sailing, hunting, skateboarding, billiards, curling, competitive eating, rock climbing, bodybuilding, horse racing, figure skating, ballroom dancing, archery, motor racing, surfing, parkour, contract bridge, crossword solving, weightlifting, synchronized swimming, birdwatching & Is $x$ a sport?; Is $x$ a game?; Is $x$ physically demanding?; Is $x$ an art form?; Is $x$ mainly a hobby?; Is $x$ dangerous? \\ \addlinespace
Foods & tomato, avocado, peanut, mushroom, rhubarb, corn, olive, cucumber, pumpkin, quinoa, coconut, eggplant, chickpea, almond, bell pepper, sweet potato, popcorn, dark chocolate, honey, seaweed, cashew, zucchini, plantain, granola, tofu, green bean, pistachio, jackfruit, oatmeal, kale & Would you call $x$ a fruit?; Would you call $x$ a vegetable?; Would you call $x$ a nut?; Is $x$ healthy?; Is $x$ sweet?; Is $x$ a snack food? \\ \addlinespace
Animals & penguin, bat, whale, ostrich, platypus, octopus, dolphin, chicken, seal, crocodile, turtle, shark, parrot, rat, pig, horse, goldfish, spider, snake, crow, raccoon, goat, jellyfish, hedgehog, wolf, hamster, pigeon, squid, ferret, cow & Is a $x$ intelligent?; Is a $x$ dangerous?; Is a $x$ a good pet?; Is a $x$ cute?; Is a $x$ graceful?; Is a $x$ a social animal? \\ \addlinespace
Social groups & politicians, lawyers, journalists, car salespeople, doctors, bankers, police officers, teachers, CEOs, social media influencers, scientists, real estate agents, professional athletes, celebrities, economists, nurses, judges, soldiers, landlords, software engineers, pharmacists, union leaders, clergy, accountants, artists, farmers, entrepreneurs, diplomats, insurance agents, university professors & Are most $x$ honest?; Are most $x$ trustworthy?; Are most $x$ overpaid?; Do most $x$ care about the public good?; Are most $x$ competent?; Are most $x$ respected? \\ \addlinespace
\bottomrule
\end{tabular}
\end{adjustbox}
\end{table}
 
Each conversation is rendered with the model's chat template, with thinking disabled for Qwen3. When a template has no system role, or places the system prompt after the first user turn, the system prompt is prepended to the first user turn so that the instruction precedes the first question. Each framing has two system prompts that differ only in the order of the answer words; every quantity is averaged over the two. In the statement framing the templates are restated as statements (e.g.\ \emph{Statement: $x$ is a sport.}) and answered True or False.
 
\begin{table}[!htbp]
\centering
\caption{System prompts. The second prompt of each framing swaps the order of the answer words (``No or Yes'', ``False or True'').}
\label{tab:prompts}
\small
\begin{adjustbox}{max width=\linewidth}
\begin{tabular}{lp{0.78\linewidth}}
\toprule
Framing & First system prompt \\
\midrule
Survey (primary) & You are a randomly selected adult taking a short opinion survey. Answer each question with a single word: Yes or No. \\
Assistant & You are a helpful assistant. Answer the user's question with a single word: Yes or No. \\
Statement & You are a randomly selected adult taking a short opinion survey. For each statement, answer with a single word: True or False. \\
\bottomrule
\end{tabular}
\end{adjustbox}
\end{table}
 
The answer is read from the first tokens of each answer word, with and without a leading space and in either case; $\ell$ is the log of the summed probability of the Yes tokens minus that of the No tokens.
 
\FloatBarrier
\section{Sample Sizes}
\label{app:samples}
 
\begin{table}[!htbp]
\centering
\caption{Pairs and general-ability samples per seed. All pairs used for evaluation come from the test half (600 pairs per seed); the profile and head ranking use the selection half.}
\label{tab:samples}
\small
\begin{adjustbox}{max width=\linewidth}
\begin{tabular}{lll}
\toprule
Measurement & Pairs & Other data \\
\midrule
Coupling, geometry $\rho$, patches and ablations at $l^*$ & 600 test & -- \\
Layer profile and readout band & 300 selection & 13 layers \\
Band removal and read gain (readout stage) & 400 / 600 test & 8 band start points \\
Head ranking & 300 selection & attribution patching \\
Head validation & 200 test & $k \in \{1,3,5,10,20,50\}$ \\
Control (Table~\ref{tab:main}) & 300 test & 500 MMLU questions, 200 passages \\
Trade-off sweep and sink test (Section~\ref{sec:control}) & 200 test & 200 MMLU questions, 100 passages \\
Confidence bins & 600 test & -- \\
Mean projection $\mu$ and entanglement statistics & -- & 100 passages \\
\bottomrule
\end{tabular}
\end{adjustbox}
\end{table}
 
MMLU questions are a fixed random sample of the \texttt{cais/mmlu} test set, scored zero-shot from the logits of the four answer letters under a helpful-assistant system prompt. Perplexity uses WikiText-2 test passages longer than 300 characters, truncated to 256 tokens.
 
\FloatBarrier
\section{Checks}
\label{app:checks}
 
For every model, framing and seed, we verify that the Yes and No conversations differ in exactly one token and share an identical prefix up to it, and run the two exactness checks of Section~\ref{sec:setup}. All 3{,}600 pairs pass in every run.
 
\begin{table}[!htbp]
\centering
\caption{Checks, summed or averaged over three seeds.}
\label{tab:checks}
\scriptsize
\begin{adjustbox}{max width=\linewidth}
\begin{tabular}{llcccc}
\toprule
Model & Framing & Pairs kept & One-token difference (\%) & Patch from embedding (target 1) & Ablate every layer (target 0) \\
\midrule
Qwen2.5-0.5B-Instruct & survey & 3600/3600 & $100.0 \pm 0.0$ & $0.998 \pm 0.005$ & $0.004 \pm 0.004$ \\
Qwen2.5-1.5B-Instruct & survey & 3600/3600 & $100.0 \pm 0.0$ & $0.997 \pm 0.001$ & $0.001 \pm 0.002$ \\
Qwen2.5-3B-Instruct & survey & 3600/3600 & $100.0 \pm 0.0$ & $0.997 \pm 0.002$ & $-0.001 \pm 0.002$ \\
Qwen2.5-7B-Instruct & assistant & 3600/3600 & $100.0 \pm 0.0$ & $1.002 \pm 0.005$ & $-0.000 \pm 0.005$ \\
Qwen2.5-7B-Instruct & statement & 3600/3600 & $100.0 \pm 0.0$ & $0.997 \pm 0.010$ & $0.001 \pm 0.001$ \\
Qwen2.5-7B-Instruct & survey & 3600/3600 & $100.0 \pm 0.0$ & $1.005 \pm 0.005$ & $-0.003 \pm 0.002$ \\
Qwen2.5-7B$^\dagger$ & survey & 3600/3600 & $100.0 \pm 0.0$ & $0.996 \pm 0.004$ & $0.000 \pm 0.006$ \\
Qwen2.5-14B-Instruct & survey & 3600/3600 & $100.0 \pm 0.0$ & $0.999 \pm 0.005$ & $0.002 \pm 0.001$ \\
Qwen2.5-14B-Instruct (4-bit) & survey & 3600/3600 & $100.0 \pm 0.0$ & $1.001 \pm 0.003$ & $-0.003 \pm 0.002$ \\
Qwen2.5-32B-Instruct & survey & 3600/3600 & $100.0 \pm 0.0$ & $1.001 \pm 0.001$ & $0.002 \pm 0.001$ \\
Qwen2.5-32B-Instruct (4-bit) & survey & 3600/3600 & $100.0 \pm 0.0$ & $0.999 \pm 0.001$ & $0.000 \pm 0.001$ \\
Qwen2.5-72B-Instruct (4-bit) & survey & 3600/3600 & $100.0 \pm 0.0$ & $1.000 \pm 0.001$ & $-0.001 \pm 0.001$ \\
Qwen3-0.6B & survey & 3600/3600 & $100.0 \pm 0.0$ & $0.998 \pm 0.001$ & $-0.000 \pm 0.001$ \\
Qwen3-1.7B & survey & 3600/3600 & $100.0 \pm 0.0$ & $1.000 \pm 0.006$ & $-0.002 \pm 0.008$ \\
Qwen3-4B & survey & 3600/3600 & $100.0 \pm 0.0$ & $0.999 \pm 0.002$ & $-0.000 \pm 0.000$ \\
Qwen3-4B-Base$^\dagger$ & survey & 3600/3600 & $100.0 \pm 0.0$ & $1.011 \pm 0.006$ & $0.000 \pm 0.002$ \\
Qwen3-4B-Instruct-2507 & assistant & 3600/3600 & $100.0 \pm 0.0$ & $1.001 \pm 0.002$ & $-0.001 \pm 0.004$ \\
Qwen3-4B-Instruct-2507 & statement & 3600/3600 & $100.0 \pm 0.0$ & $0.999 \pm 0.004$ & $0.002 \pm 0.001$ \\
Qwen3-4B-Instruct-2507 & survey & 3600/3600 & $100.0 \pm 0.0$ & $0.999 \pm 0.002$ & $-0.000 \pm 0.001$ \\
Qwen3-8B & survey & 3600/3600 & $100.0 \pm 0.0$ & $1.000 \pm 0.002$ & $0.001 \pm 0.001$ \\
Qwen3-14B & survey & 3600/3600 & $100.0 \pm 0.0$ & $0.997 \pm 0.003$ & $0.000 \pm 0.003$ \\
Qwen3-14B (4-bit) & survey & 3600/3600 & $100.0 \pm 0.0$ & $1.000 \pm 0.000$ & $-0.001 \pm 0.002$ \\
Qwen3-32B & survey & 3600/3600 & $100.0 \pm 0.0$ & $1.001 \pm 0.004$ & $0.002 \pm 0.001$ \\
Qwen3-32B (4-bit) & survey & 3600/3600 & $100.0 \pm 0.0$ & $0.997 \pm 0.000$ & $-0.001 \pm 0.002$ \\
Llama-3.1-8B-Instruct & assistant & 3600/3600 & $100.0 \pm 0.0$ & $1.001 \pm 0.001$ & $-0.000 \pm 0.000$ \\
Llama-3.1-8B-Instruct & statement & 3600/3600 & $100.0 \pm 0.0$ & $0.998 \pm 0.002$ & $0.001 \pm 0.002$ \\
Llama-3.1-8B-Instruct & survey & 3600/3600 & $100.0 \pm 0.0$ & $0.998 \pm 0.001$ & $-0.000 \pm 0.000$ \\
Llama-3.1-70B-Instruct (4-bit) & survey & 3600/3600 & $100.0 \pm 0.0$ & $1.000 \pm 0.001$ & $-0.000 \pm 0.002$ \\
Gemma-2-9B-it & assistant & 3600/3600 & $100.0 \pm 0.0$ & $0.999 \pm 0.001$ & $-0.001 \pm 0.001$ \\
Gemma-2-9B-it & statement & 3600/3600 & $100.0 \pm 0.0$ & $1.000 \pm 0.001$ & $-0.000 \pm 0.001$ \\
Gemma-2-9B-it & survey & 3600/3600 & $100.0 \pm 0.0$ & $0.999 \pm 0.002$ & $-0.000 \pm 0.001$ \\
Gemma-2-27B-it & survey & 3600/3600 & $100.0 \pm 0.0$ & $0.999 \pm 0.001$ & $0.001 \pm 0.002$ \\
Gemma-2-27B-it (4-bit) & assistant & 3600/3600 & $100.0 \pm 0.0$ & $1.001 \pm 0.001$ & $-0.000 \pm 0.000$ \\
Gemma-2-27B-it (4-bit) & statement & 3600/3600 & $100.0 \pm 0.0$ & $1.000 \pm 0.002$ & $-0.003 \pm 0.001$ \\
Gemma-2-27B-it (4-bit) & survey & 3600/3600 & $100.0 \pm 0.0$ & $0.999 \pm 0.001$ & $-0.000 \pm 0.003$ \\
Mistral-7B-Instruct-v0.3 & assistant & 3600/3600 & $100.0 \pm 0.0$ & $1.000 \pm 0.003$ & $0.000 \pm 0.000$ \\
Mistral-7B-Instruct-v0.3 & statement & 3600/3600 & $100.0 \pm 0.0$ & $0.997 \pm 0.001$ & $-0.000 \pm 0.001$ \\
Mistral-7B-Instruct-v0.3 & survey & 3600/3600 & $100.0 \pm 0.0$ & $0.999 \pm 0.001$ & $0.000 \pm 0.001$ \\
Mistral-7B-v0.3$^\dagger$ & survey & 3600/3600 & $100.0 \pm 0.0$ & $0.999 \pm 0.002$ & $-0.001 \pm 0.001$ \\
\bottomrule
\end{tabular}
\end{adjustbox}
\end{table}
 
\FloatBarrier
\section{Per-Model Results}
\label{app:fullresults}
 
Tables~\ref{tab:full1} and~\ref{tab:full2} give the values summarised in Table~\ref{tab:main} with standard deviations over three seeds, together with the selected layer and readout band. Rows marked 4-bit below a bfloat16 row are the quantisation checks.
 
\begin{table}[!htbp]
\centering
\caption{Behaviour, geometry and edits at $l^*$ (primary framing, held-out pairs). Patch, shared and specific: share carried by adding $d$, its component along the leave-one-domain-out $g$, and the remainder, at the first answer token at $l^*$. Ablate $g$ at $l^*$: share remaining after ablating $g$ at that token and layer only.}
\label{tab:full1}
\scriptsize
\setlength{\tabcolsep}{3pt}
\begin{adjustbox}{max width=\linewidth}
\begin{tabular}{lccccccccc}
\toprule
Model & $|\Delta P|$ & Pos.\ (\%) & $\rho$ & Partial $\rho$ & Random weights $\rho$ & Patch & Shared & Specific & Ablate $g$ at $l^*$ \\
\midrule
Qwen2.5-0.5B-Instruct & $0.144 \pm 0.002$ & $70.6 \pm 1.1$ & $0.56 \pm 0.05$ & $0.50 \pm 0.04$ & $-0.01 \pm 0.05$ & $0.83 \pm 0.01$ & $0.81 \pm 0.02$ & $0.07 \pm 0.01$ & $0.18 \pm 0.03$ \\
Qwen2.5-1.5B-Instruct & $0.216 \pm 0.012$ & $18.4 \pm 0.6$ & $0.70 \pm 0.03$ & $0.49 \pm 0.10$ & $-0.02 \pm 0.04$ & $0.34 \pm 0.01$ & $0.34 \pm 0.01$ & $-0.00 \pm 0.00$ & $0.64 \pm 0.01$ \\
Qwen2.5-3B-Instruct & $0.132 \pm 0.011$ & $32.8 \pm 2.0$ & $0.73 \pm 0.02$ & $0.36 \pm 0.09$ & $-0.31 \pm 0.08$ & $0.55 \pm 0.48$ & $0.42 \pm 0.37$ & $0.10 \pm 0.09$ & $0.55 \pm 0.39$ \\
Qwen2.5-7B-Instruct & $0.087 \pm 0.018$ & $44.3 \pm 2.4$ & $0.72 \pm 0.04$ & $0.45 \pm 0.08$ & $-0.00 \pm 0.07$ & $0.73 \pm 0.08$ & $0.56 \pm 0.05$ & $0.16 \pm 0.01$ & $0.36 \pm 0.04$ \\
Qwen2.5-7B$^\dagger$ & $0.091 \pm 0.001$ & $62.9 \pm 2.5$ & $0.61 \pm 0.07$ & $0.52 \pm 0.08$ & -- & $0.91 \pm 0.03$ & $0.82 \pm 0.02$ & $0.09 \pm 0.01$ & $0.17 \pm 0.02$ \\
Qwen2.5-14B-Instruct & $0.118 \pm 0.025$ & $15.9 \pm 0.3$ & $0.53 \pm 0.04$ & $0.27 \pm 0.08$ & -- & $0.62 \pm 0.02$ & $0.53 \pm 0.03$ & $0.03 \pm 0.00$ & $0.41 \pm 0.02$ \\
Qwen2.5-14B-Instruct (4-bit) & $0.134 \pm 0.025$ & $8.4 \pm 1.0$ & $0.44 \pm 0.03$ & $0.22 \pm 0.04$ & -- & $0.78 \pm 0.02$ & $0.72 \pm 0.03$ & $0.03 \pm 0.00$ & $0.30 \pm 0.02$ \\
Qwen2.5-32B-Instruct & $0.075 \pm 0.013$ & $20.2 \pm 0.2$ & $0.74 \pm 0.06$ & $0.54 \pm 0.07$ & -- & $0.69 \pm 0.07$ & $0.55 \pm 0.09$ & $0.07 \pm 0.01$ & $0.39 \pm 0.05$ \\
Qwen2.5-32B-Instruct (4-bit) & $0.069 \pm 0.014$ & $19.6 \pm 0.3$ & $0.75 \pm 0.05$ & $0.54 \pm 0.06$ & -- & $0.65 \pm 0.07$ & $0.55 \pm 0.08$ & $0.06 \pm 0.01$ & $0.41 \pm 0.05$ \\
Qwen2.5-72B-Instruct (4-bit) & $0.117 \pm 0.022$ & $35.7 \pm 0.9$ & $0.81 \pm 0.04$ & $0.66 \pm 0.05$ & -- & $0.72 \pm 0.12$ & $0.61 \pm 0.11$ & $-0.03 \pm 0.01$ & $0.54 \pm 0.05$ \\
Qwen3-0.6B & $0.707 \pm 0.001$ & $100.0 \pm 0.0$ & $0.43 \pm 0.04$ & $0.27 \pm 0.03$ & $-0.13 \pm 0.08$ & $0.39 \pm 0.20$ & $0.39 \pm 0.20$ & $0.01 \pm 0.01$ & $0.55 \pm 0.23$ \\
Qwen3-1.7B & $0.089 \pm 0.005$ & $44.0 \pm 1.5$ & $0.50 \pm 0.05$ & $0.45 \pm 0.03$ & $-0.06 \pm 0.04$ & $0.53 \pm 0.04$ & $0.45 \pm 0.04$ & $0.01 \pm 0.01$ & $0.46 \pm 0.03$ \\
Qwen3-4B & $0.171 \pm 0.020$ & $51.6 \pm 0.6$ & $0.72 \pm 0.02$ & $0.64 \pm 0.09$ & $0.01 \pm 0.03$ & $0.50 \pm 0.11$ & $0.48 \pm 0.11$ & $0.02 \pm 0.04$ & $0.51 \pm 0.06$ \\
Qwen3-4B-Base$^\dagger$ & $0.068 \pm 0.002$ & $74.1 \pm 2.1$ & $0.45 \pm 0.05$ & $0.51 \pm 0.04$ & -- & $1.21 \pm 0.03$ & $1.17 \pm 0.01$ & $0.07 \pm 0.01$ & $-0.15 \pm 0.01$ \\
Qwen3-4B-Instruct-2507 & $0.169 \pm 0.024$ & $11.7 \pm 1.3$ & $0.54 \pm 0.02$ & $0.54 \pm 0.01$ & $0.19 \pm 0.05$ & $0.87 \pm 0.01$ & $0.78 \pm 0.02$ & $0.00 \pm 0.01$ & $0.16 \pm 0.01$ \\
Qwen3-8B & $0.086 \pm 0.010$ & $44.8 \pm 0.9$ & $0.64 \pm 0.08$ & $0.55 \pm 0.08$ & $0.12 \pm 0.08$ & $0.49 \pm 0.09$ & $0.31 \pm 0.13$ & $0.10 \pm 0.04$ & $0.60 \pm 0.11$ \\
Qwen3-14B & $0.078 \pm 0.013$ & $41.1 \pm 1.9$ & $0.77 \pm 0.03$ & $0.68 \pm 0.05$ & -- & $0.32 \pm 0.16$ & $0.21 \pm 0.14$ & $0.05 \pm 0.03$ & $0.75 \pm 0.10$ \\
Qwen3-14B (4-bit) & $0.096 \pm 0.016$ & $35.1 \pm 1.2$ & $0.74 \pm 0.03$ & $0.75 \pm 0.03$ & -- & $0.40 \pm 0.02$ & $0.31 \pm 0.02$ & $0.08 \pm 0.01$ & $0.71 \pm 0.02$ \\
Qwen3-32B & $0.085 \pm 0.006$ & $47.0 \pm 1.2$ & $0.83 \pm 0.03$ & $0.61 \pm 0.07$ & -- & $0.13 \pm 0.03$ & $0.08 \pm 0.02$ & $-0.01 \pm 0.03$ & $0.69 \pm 0.01$ \\
Qwen3-32B (4-bit) & $0.080 \pm 0.007$ & $45.5 \pm 2.1$ & $0.84 \pm 0.02$ & $0.66 \pm 0.08$ & -- & $0.14 \pm 0.03$ & $0.09 \pm 0.02$ & $0.00 \pm 0.02$ & $0.69 \pm 0.01$ \\
Llama-3.1-8B-Instruct & $0.171 \pm 0.012$ & $18.6 \pm 0.9$ & $0.70 \pm 0.06$ & $0.56 \pm 0.04$ & $-0.06 \pm 0.03$ & $0.41 \pm 0.02$ & $0.32 \pm 0.02$ & $0.06 \pm 0.00$ & $0.63 \pm 0.01$ \\
Llama-3.1-70B-Instruct (4-bit) & $0.059 \pm 0.012$ & $36.0 \pm 0.5$ & $0.85 \pm 0.02$ & $0.63 \pm 0.07$ & -- & $0.41 \pm 0.17$ & $0.28 \pm 0.14$ & $0.01 \pm 0.06$ & $0.52 \pm 0.09$ \\
Gemma-2-9B-it & $0.056 \pm 0.019$ & $55.9 \pm 1.1$ & $0.80 \pm 0.07$ & $0.55 \pm 0.01$ & $0.08 \pm 0.09$ & $0.38 \pm 0.30$ & $0.26 \pm 0.23$ & $0.06 \pm 0.04$ & $0.67 \pm 0.29$ \\
Gemma-2-27B-it & $0.071 \pm 0.020$ & $47.1 \pm 1.5$ & $0.79 \pm 0.04$ & $0.38 \pm 0.03$ & -- & $0.25 \pm 0.12$ & $0.12 \pm 0.14$ & $0.15 \pm 0.08$ & $0.88 \pm 0.17$ \\
Gemma-2-27B-it (4-bit) & $0.077 \pm 0.018$ & $46.7 \pm 0.5$ & $0.80 \pm 0.05$ & $0.37 \pm 0.01$ & -- & $0.30 \pm 0.14$ & $0.16 \pm 0.19$ & $0.17 \pm 0.09$ & $0.85 \pm 0.19$ \\
Mistral-7B-Instruct-v0.3 & $0.087 \pm 0.022$ & $48.2 \pm 1.0$ & $0.81 \pm 0.05$ & $0.43 \pm 0.03$ & $-0.05 \pm 0.02$ & $0.35 \pm 0.02$ & $0.25 \pm 0.01$ & $0.01 \pm 0.01$ & $0.66 \pm 0.03$ \\
Mistral-7B-v0.3$^\dagger$ & $0.152 \pm 0.003$ & $93.3 \pm 0.9$ & $0.61 \pm 0.08$ & $0.46 \pm 0.09$ & -- & $1.01 \pm 0.03$ & $0.99 \pm 0.03$ & $0.01 \pm 0.01$ & $0.05 \pm 0.05$ \\
\bottomrule
\end{tabular}
\end{adjustbox}
\end{table}
 
\begin{table}[!htbp]
\centering
\caption{Heads, band ablation, reader and layers. Heads: share carried by patching the top-10 or 10 random heads, and share remaining after knocking out the top 10. $\kappa_g$, $\kappa_\text{rand}$: control-stage band ablation at every position (base models: from the 200-pair sink test, marked $\ddagger$). Read neg.: questions with negative read gain (\%). Read, write: within-model Spearman $\rho$ of the pair's read gain and of the first question's write strength with coupling; LOO: read gain of $q_2$ from its other first questions. $l^*$ and band: layer indices (hidden-state index, 0 = embedding), mean over seeds and range of band ends. Dashes: not measured.}
\label{tab:full2}
\scriptsize
\setlength{\tabcolsep}{2.5pt}
\begin{adjustbox}{max width=\linewidth}
\begin{tabular}{lcccccccccccc}
\toprule
Model & Top-10 & Random-10 & Knockout & $\kappa_g$ & $\kappa_\text{rand}$ & Read neg. & Read & Write & LOO & $L$ & $l^*$ & Band \\
\midrule
Qwen2.5-0.5B-Instruct & $1.07 \pm 0.03$ & $0.10 \pm 0.43$ & $-0.17 \pm 0.05$ & $-0.00 \pm 0.01$ & $0.75 \pm 0.08$ & $27.5 \pm 0.8$ & $0.95 \pm 0.01$ & $-0.06 \pm 0.02$ & $0.71 \pm 0.04$ & 24 & 13.0 & 9--19 \\
Qwen2.5-1.5B-Instruct & $1.44 \pm 0.11$ & $-0.01 \pm 0.02$ & $-0.36 \pm 0.11$ & $0.08 \pm 0.01$ & $0.87 \pm 0.03$ & $78.6 \pm 7.6$ & $0.58 \pm 0.06$ & $0.29 \pm 0.03$ & $0.43 \pm 0.10$ & 28 & 21.0 & 13--27 \\
Qwen2.5-3B-Instruct & $0.70 \pm 0.02$ & $0.01 \pm 0.02$ & $0.26 \pm 0.03$ & $0.10 \pm 0.06$ & $0.95 \pm 0.07$ & $70.8 \pm 20.4$ & $0.40 \pm 0.37$ & $-0.05 \pm 0.04$ & $0.27 \pm 0.26$ & 36 & 28.3 & 20/23--32 \\
Qwen2.5-7B-Instruct & $0.41 \pm 0.07$ & $0.00 \pm 0.00$ & $0.50 \pm 0.10$ & $0.13 \pm 0.03$ & $0.99 \pm 0.01$ & $42.8 \pm 2.9$ & $0.83 \pm 0.03$ & $-0.07 \pm 0.03$ & $0.33 \pm 0.10$ & 28 & 20.3 & 15--27 \\
Qwen2.5-7B$^\dagger$ & $1.06 \pm 0.15$ & $-0.01 \pm 0.01$ & $-0.15 \pm 0.12$ & $0.15 \pm 0.02$$^\ddagger$ & -- & $20.8 \pm 3.0$ & $0.95 \pm 0.00$ & $0.08 \pm 0.08$ & $0.39 \pm 0.06$ & 28 & 15.3 & 15--25 \\
Qwen2.5-14B-Instruct & $0.61 \pm 0.03$ & $0.00 \pm 0.00$ & $0.32 \pm 0.05$ & $0.06 \pm 0.03$ & $0.99 \pm 0.06$ & $88.9 \pm 3.5$ & $0.53 \pm 0.02$ & $-0.08 \pm 0.04$ & $0.40 \pm 0.03$ & 48 & 30.0 & 18/22--39 \\
Qwen2.5-14B-Instruct (4-bit) & $0.64 \pm 0.02$ & $0.00 \pm 0.00$ & $0.25 \pm 0.01$ & $0.03 \pm 0.01$ & $1.00 \pm 0.06$ & $94.4 \pm 1.7$ & $0.62 \pm 0.02$ & $-0.08 \pm 0.05$ & $0.54 \pm 0.02$ & 48 & 30.0 & 22/26--39/43 \\
Qwen2.5-32B-Instruct & $0.58 \pm 0.06$ & $0.02 \pm 0.01$ & $0.35 \pm 0.05$ & $0.15 \pm 0.05$ & $0.98 \pm 0.02$ & $94.4 \pm 2.9$ & $0.67 \pm 0.01$ & $-0.12 \pm 0.01$ & $0.21 \pm 0.04$ & 64 & 46.7 & 40--57 \\
Qwen2.5-32B-Instruct (4-bit) & $0.59 \pm 0.05$ & $0.02 \pm 0.00$ & $0.35 \pm 0.05$ & $0.16 \pm 0.03$ & $0.98 \pm 0.02$ & $96.1 \pm 2.4$ & $0.64 \pm 0.01$ & $-0.12 \pm 0.05$ & $0.16 \pm 0.07$ & 64 & 46.7 & 40--57 \\
Qwen2.5-72B-Instruct (4-bit) & $0.28 \pm 0.01$ & $-0.00 \pm 0.01$ & $0.75 \pm 0.02$ & $0.10 \pm 0.04$ & $1.00 \pm 0.01$ & -- & -- & -- & -- & 80 & 55.0 & 44--72/79 \\
Qwen3-0.6B & $1.06 \pm 0.01$ & $0.04 \pm 0.05$ & $-0.18 \pm 0.00$ & $0.06 \pm 0.00$ & $1.02 \pm 0.05$ & $0.0 \pm 0.0$ & $0.64 \pm 0.17$ & $0.27 \pm 0.32$ & $0.65 \pm 0.06$ & 28 & 21.3 & 15--27 \\
Qwen3-1.7B & $0.61 \pm 0.01$ & $0.01 \pm 0.01$ & $0.36 \pm 0.02$ & $0.07 \pm 0.00$ & $0.94 \pm 0.16$ & $40.6 \pm 5.7$ & $0.75 \pm 0.03$ & $-0.12 \pm 0.07$ & $0.42 \pm 0.02$ & 28 & 18.0 & 13--25 \\
Qwen3-4B & $0.69 \pm 0.02$ & $0.04 \pm 0.04$ & $0.30 \pm 0.02$ & $0.03 \pm 0.00$ & $0.93 \pm 0.04$ & $13.9 \pm 9.3$ & $0.64 \pm 0.08$ & $0.08 \pm 0.13$ & $0.50 \pm 0.06$ & 36 & 21.7 & 10--32 \\
Qwen3-4B-Base$^\dagger$ & $1.34 \pm 0.17$ & $0.03 \pm 0.02$ & $-0.31 \pm 0.18$ & $-0.15 \pm 0.15$$^\ddagger$ & -- & $0.0 \pm 0.0$ & $0.70 \pm 0.01$ & $-0.09 \pm 0.02$ & $0.42 \pm 0.03$ & 36 & 20.0 & 29--35 \\
Qwen3-4B-Instruct-2507 & $1.24 \pm 0.03$ & $-0.01 \pm 0.03$ & $-0.27 \pm 0.04$ & $0.03 \pm 0.01$ & $1.06 \pm 0.08$ & $95.8 \pm 2.2$ & $0.83 \pm 0.03$ & $0.15 \pm 0.06$ & $0.59 \pm 0.04$ & 36 & 17.7 & 13--29 \\
Qwen3-8B & $0.48 \pm 0.03$ & $0.04 \pm 0.02$ & $0.44 \pm 0.03$ & $0.12 \pm 0.03$ & $1.00 \pm 0.01$ & $25.0 \pm 11.0$ & $0.67 \pm 0.15$ & $0.14 \pm 0.06$ & $0.56 \pm 0.04$ & 36 & 21.3 & 13/16--26/32 \\
Qwen3-14B & $0.59 \pm 0.02$ & $0.01 \pm 0.01$ & $0.40 \pm 0.01$ & $0.09 \pm 0.01$ & $1.02 \pm 0.03$ & $59.2 \pm 2.9$ & $0.58 \pm 0.20$ & $-0.05 \pm 0.06$ & $0.35 \pm 0.11$ & 40 & 24.0 & 15--29 \\
Qwen3-14B (4-bit) & $0.64 \pm 0.01$ & $0.01 \pm 0.00$ & $0.36 \pm 0.02$ & $0.11 \pm 0.02$ & $1.02 \pm 0.02$ & $75.8 \pm 1.4$ & $0.76 \pm 0.01$ & $0.01 \pm 0.05$ & $0.41 \pm 0.02$ & 40 & 22.0 & 15--29 \\
Qwen3-32B & $0.41 \pm 0.02$ & $0.00 \pm 0.01$ & $0.56 \pm 0.03$ & $0.07 \pm 0.03$ & $1.06 \pm 0.01$ & $66.9 \pm 12.4$ & $0.12 \pm 0.04$ & $0.27 \pm 0.06$ & $0.18 \pm 0.03$ & 64 & 51.0 & 35--52/57 \\
Qwen3-32B (4-bit) & $0.40 \pm 0.01$ & $0.00 \pm 0.00$ & $0.58 \pm 0.04$ & $0.07 \pm 0.03$ & $1.02 \pm 0.02$ & $81.7 \pm 10.1$ & $0.08 \pm 0.06$ & $0.27 \pm 0.07$ & $0.12 \pm 0.05$ & 64 & 50.7 & 35--52/57 \\
Llama-3.1-8B-Instruct & $0.79 \pm 0.01$ & $-0.01 \pm 0.01$ & $0.15 \pm 0.02$ & $0.02 \pm 0.00$ & $0.98 \pm 0.06$ & $87.8 \pm 0.5$ & $0.61 \pm 0.02$ & $0.09 \pm 0.02$ & $0.45 \pm 0.06$ & 32 & 15.0 & 4--20 \\
Llama-3.1-70B-Instruct (4-bit) & $0.40 \pm 0.01$ & $-0.00 \pm 0.01$ & $0.56 \pm 0.01$ & $0.10 \pm 0.05$ & $1.01 \pm 0.06$ & -- & -- & -- & -- & 80 & 34.7 & 15/22--79 \\
Gemma-2-9B-it & $0.42 \pm 0.02$ & $0.02 \pm 0.03$ & $0.55 \pm 0.02$ & $0.03 \pm 0.01$ & $1.01 \pm 0.02$ & $32.8 \pm 35.2$ & $0.51 \pm 0.32$ & $0.10 \pm 0.14$ & $0.28 \pm 0.14$ & 42 & 30.3 & 8/12--37/41 \\
Gemma-2-27B-it & $0.48 \pm 0.03$ & $0.01 \pm 0.02$ & $0.53 \pm 0.01$ & $-0.00 \pm 0.00$ & $0.94 \pm 0.06$ & $41.9 \pm 4.3$ & $0.49 \pm 0.20$ & $0.16 \pm 0.21$ & $0.17 \pm 0.06$ & 46 & 37.7 & 5/9--45 \\
Gemma-2-27B-it (4-bit) & $0.50 \pm 0.03$ & $0.01 \pm 0.02$ & $0.50 \pm 0.02$ & $0.00 \pm 0.00$ & $0.86 \pm 0.03$ & $44.7 \pm 3.9$ & $0.51 \pm 0.24$ & $0.12 \pm 0.24$ & $0.21 \pm 0.07$ & 46 & 37.7 & 9--45 \\
Mistral-7B-Instruct-v0.3 & $0.52 \pm 0.05$ & $0.02 \pm 0.05$ & $0.42 \pm 0.04$ & $-0.03 \pm 0.01$ & $1.00 \pm 0.04$ & $49.2 \pm 1.4$ & $0.61 \pm 0.03$ & $0.16 \pm 0.02$ & $0.37 \pm 0.01$ & 32 & 17.7 & 6--23/31 \\
Mistral-7B-v0.3$^\dagger$ & $1.24 \pm 0.03$ & $0.02 \pm 0.04$ & $-0.19 \pm 0.06$ & $0.10 \pm 0.14$$^\ddagger$ & -- & $0.0 \pm 0.0$ & $0.71 \pm 0.03$ & $-0.14 \pm 0.08$ & $0.47 \pm 0.06$ & 32 & 18.0 & 6/17--31 \\
\bottomrule
\end{tabular}
\end{adjustbox}
\end{table}
 
\FloatBarrier
\section{Confidence Bins}
\label{app:confbins}
 
\begin{table}[!htbp]
\centering
\caption{Held-out pairs (pooled over seeds) by the model's standalone probability $p$ of its answer to the second question. For each bin: number of pairs, geometry $\rho$, and share remaining after band ablation of $g$ and of a random direction. Bins with fewer than 10 pairs in a seed are left blank.}
\label{tab:confbins_full}
\scriptsize
\setlength{\tabcolsep}{2.5pt}
\begin{adjustbox}{max width=\linewidth}
\begin{tabular}{lcccccccccccc}
\toprule
 & \multicolumn{4}{c}{$p < 0.9$} & \multicolumn{4}{c}{$0.9 \le p < 0.99$} & \multicolumn{4}{c}{$p \ge 0.99$} \\
\cmidrule(lr){2-5}\cmidrule(lr){6-9}\cmidrule(lr){10-13}
Model & $n$ & $\rho$ & $\kappa_g$ & $\kappa_\text{rand}$ & $n$ & $\rho$ & $\kappa_g$ & $\kappa_\text{rand}$ & $n$ & $\rho$ & $\kappa_g$ & $\kappa_\text{rand}$ \\
\midrule
Qwen2.5-0.5B-Instruct & 1590 & $0.54$ & $0.00$ & $0.94$ & 210 & $0.72$ & $-0.01$ & $1.04$ & 0 & -- & -- & -- \\
Qwen2.5-1.5B-Instruct & 360 & $0.59$ & $0.08$ & $1.08$ & 435 & $0.65$ & $0.07$ & $1.08$ & 1005 & $0.74$ & $0.07$ & $1.06$ \\
Qwen2.5-3B-Instruct & 155 & $0.58$ & $0.09$ & $1.01$ & 180 & $0.76$ & $0.14$ & $0.99$ & 1465 & $0.73$ & $0.10$ & $1.00$ \\
Qwen2.5-7B-Instruct & 135 & $0.72$ & $0.07$ & $1.01$ & 195 & $0.50$ & $0.08$ & $1.00$ & 1470 & $0.74$ & $0.14$ & $0.99$ \\
Qwen2.5-14B-Instruct & 120 & $0.38$ & $-0.01$ & $1.01$ & 130 & $0.44$ & $0.04$ & $1.01$ & 1550 & $0.55$ & $0.06$ & $1.01$ \\
Qwen2.5-32B-Instruct & 100 & $0.39$ & $0.10$ & $1.00$ & 55 & $0.18$ & $0.07$ & $1.02$ & 1645 & $0.77$ & $0.15$ & $1.01$ \\
Qwen2.5-32B-Instruct (4-bit) & 90 & $0.17$ & $0.06$ & $1.00$ & 100 & $0.44$ & $0.09$ & $1.00$ & 1610 & $0.78$ & $0.16$ & $1.00$ \\
Qwen2.5-72B-Instruct (4-bit) & 210 & $0.61$ & $0.16$ & $0.97$ & 215 & $0.59$ & $0.15$ & $0.97$ & 1375 & $0.83$ & $0.09$ & $0.97$ \\
Qwen3-0.6B & 1795 & $0.44$ & $0.06$ & $1.20$ & 5 & -- & -- & -- & 0 & -- & -- & -- \\
Qwen3-1.7B & 240 & $0.43$ & $0.07$ & $0.95$ & 240 & $0.39$ & $0.08$ & $0.97$ & 1320 & $0.54$ & $0.06$ & $0.97$ \\
Qwen3-4B & 205 & $0.69$ & $0.03$ & $0.96$ & 170 & $0.54$ & $0.05$ & $0.96$ & 1425 & $0.74$ & $0.03$ & $0.94$ \\
Qwen3-4B-Instruct-2507 & 95 & $0.49$ & $0.01$ & $0.95$ & 130 & $0.37$ & $0.01$ & $0.96$ & 1575 & $0.56$ & $0.02$ & $0.99$ \\
Qwen3-8B & 175 & $0.50$ & $0.14$ & $1.01$ & 155 & $0.49$ & $0.16$ & $1.01$ & 1470 & $0.67$ & $0.11$ & $1.03$ \\
Qwen3-14B & 85 & $0.76$ & $0.08$ & $0.93$ & 85 & $0.65$ & $0.08$ & $0.94$ & 1630 & $0.77$ & $0.09$ & $0.97$ \\
Qwen3-32B & 410 & $0.40$ & $0.04$ & $0.95$ & 500 & $0.76$ & $0.08$ & $0.98$ & 890 & $0.90$ & $0.07$ & $1.00$ \\
Qwen3-32B (4-bit) & 435 & $0.52$ & $0.04$ & $0.95$ & 425 & $0.77$ & $0.08$ & $0.96$ & 940 & $0.91$ & $0.07$ & $0.97$ \\
Llama-3.1-8B-Instruct & 550 & $0.61$ & $0.03$ & $1.01$ & 490 & $0.69$ & $0.02$ & $1.00$ & 760 & $0.80$ & $0.01$ & $0.99$ \\
Llama-3.1-70B-Instruct (4-bit) & 365 & $0.61$ & $0.17$ & $1.00$ & 335 & $0.76$ & $0.12$ & $0.98$ & 1100 & $0.89$ & $0.08$ & $0.97$ \\
Gemma-2-9B-it & 155 & $0.42$ & $0.04$ & $1.00$ & 210 & $0.64$ & $0.05$ & $0.99$ & 1435 & $0.82$ & $0.03$ & $0.99$ \\
Gemma-2-27B-it & 115 & $0.32$ & $0.01$ & $0.92$ & 265 & $0.35$ & $-0.00$ & $0.91$ & 1420 & $0.85$ & $-0.00$ & $0.94$ \\
Gemma-2-27B-it (4-bit) & 165 & $0.43$ & $0.00$ & $0.90$ & 245 & $0.39$ & $0.01$ & $0.85$ & 1390 & $0.86$ & $-0.00$ & $0.87$ \\
Mistral-7B-Instruct-v0.3 & 150 & $0.52$ & $-0.01$ & $1.04$ & 135 & $0.71$ & $-0.04$ & $1.02$ & 1515 & $0.83$ & $-0.02$ & $0.99$ \\
\bottomrule
\end{tabular}
\end{adjustbox}
\end{table}
 
\FloatBarrier
\section{Quantisation Check}
\label{app:4bit}
 
\begin{table}[!htbp]
\centering
\caption{bfloat16 and 4-bit NF4 runs of the same five models (primary framing; means $\pm$ s.d.\ over seeds). PPL: perplexity ratio under control-stage band ablation of $g$.}
\label{tab:4bit}
\scriptsize
\begin{adjustbox}{max width=\linewidth}
\begin{tabular}{llcccccc}
\toprule
Model & Precision & Pos.\ (\%) & $\rho$ & Patch & Top-10 heads & $\kappa_g$ & PPL \\
\midrule
Qwen2.5-14B-Instruct & bf16 & $15.9 \pm 0.3$ & $0.53 \pm 0.04$ & $0.62 \pm 0.02$ & $0.61 \pm 0.03$ & $0.06 \pm 0.03$ & $1.074 \pm 0.017$ \\
Qwen2.5-14B-Instruct & 4-bit & $8.4 \pm 1.0$ & $0.44 \pm 0.03$ & $0.78 \pm 0.02$ & $0.64 \pm 0.02$ & $0.03 \pm 0.01$ & $1.128 \pm 0.071$ \\
Qwen2.5-32B-Instruct & bf16 & $20.2 \pm 0.2$ & $0.74 \pm 0.06$ & $0.69 \pm 0.07$ & $0.58 \pm 0.06$ & $0.15 \pm 0.05$ & $1.026 \pm 0.005$ \\
Qwen2.5-32B-Instruct & 4-bit & $19.6 \pm 0.3$ & $0.75 \pm 0.05$ & $0.65 \pm 0.07$ & $0.59 \pm 0.05$ & $0.16 \pm 0.03$ & $1.018 \pm 0.007$ \\
Qwen3-14B & bf16 & $41.1 \pm 1.9$ & $0.77 \pm 0.03$ & $0.32 \pm 0.16$ & $0.59 \pm 0.02$ & $0.09 \pm 0.01$ & $1.050 \pm 0.001$ \\
Qwen3-14B & 4-bit & $35.1 \pm 1.2$ & $0.74 \pm 0.03$ & $0.40 \pm 0.02$ & $0.64 \pm 0.01$ & $0.11 \pm 0.02$ & $1.041 \pm 0.002$ \\
Qwen3-32B & bf16 & $47.0 \pm 1.2$ & $0.83 \pm 0.03$ & $0.13 \pm 0.03$ & $0.41 \pm 0.02$ & $0.07 \pm 0.03$ & $1.785 \pm 0.374$ \\
Qwen3-32B & 4-bit & $45.5 \pm 2.1$ & $0.84 \pm 0.02$ & $0.14 \pm 0.03$ & $0.40 \pm 0.01$ & $0.07 \pm 0.03$ & $1.784 \pm 0.367$ \\
Gemma-2-27B-it & bf16 & $47.1 \pm 1.5$ & $0.79 \pm 0.04$ & $0.25 \pm 0.12$ & $0.48 \pm 0.03$ & $-0.00 \pm 0.00$ & $13.927 \pm 2.532$ \\
Gemma-2-27B-it & 4-bit & $46.7 \pm 0.5$ & $0.80 \pm 0.05$ & $0.30 \pm 0.14$ & $0.50 \pm 0.03$ & $0.00 \pm 0.00$ & $14.789 \pm 1.505$ \\
\bottomrule
\end{tabular}
\end{adjustbox}
\end{table}
 
\FloatBarrier
\section{Framings and Personas}
\label{app:persona}
 
Six models were also run under the assistant and statement framings (behaviour and causal stages). The persona comparison re-ranks heads under the assistant framing on the selection half and correlates the per-head map with that of the survey framing.
 
\begin{table}[!htbp]
\centering
\caption{Robustness across framings (held-out pairs; means $\pm$ s.d.\ over seeds). Ablate $g$ at $l^*$: share remaining after ablating $g$ at the first answer token at $l^*$.}
\label{tab:framings}
\scriptsize
\begin{adjustbox}{max width=\linewidth}
\begin{tabular}{llcccccc}
\toprule
Model & Framing & Pos.\ (\%) & $\rho$ & Patch & Shared & Ablate $g$ at $l^*$ & Random at $l^*$ \\
\midrule
Qwen2.5-7B-Instruct & assistant & $61.7 \pm 1.5$ & $0.74 \pm 0.01$ & $0.55 \pm 0.41$ & $0.45 \pm 0.34$ & $0.53 \pm 0.35$ & $1.00 \pm 0.00$ \\
Qwen2.5-7B-Instruct & statement & $45.8 \pm 0.8$ & $0.73 \pm 0.04$ & $0.76 \pm 0.07$ & $0.68 \pm 0.04$ & $0.31 \pm 0.01$ & $0.99 \pm 0.00$ \\
Qwen2.5-7B-Instruct & survey & $44.3 \pm 2.4$ & $0.72 \pm 0.04$ & $0.73 \pm 0.08$ & $0.56 \pm 0.05$ & $0.36 \pm 0.04$ & $1.00 \pm 0.00$ \\
Qwen3-4B-Instruct-2507 & assistant & $81.3 \pm 1.4$ & $0.60 \pm 0.04$ & $0.41 \pm 0.35$ & $0.43 \pm 0.37$ & $0.61 \pm 0.31$ & $0.99 \pm 0.00$ \\
Qwen3-4B-Instruct-2507 & statement & $17.8 \pm 0.8$ & $0.42 \pm 0.03$ & $0.80 \pm 0.06$ & $0.74 \pm 0.11$ & $0.22 \pm 0.09$ & $1.00 \pm 0.00$ \\
Qwen3-4B-Instruct-2507 & survey & $11.7 \pm 1.3$ & $0.54 \pm 0.02$ & $0.87 \pm 0.01$ & $0.78 \pm 0.02$ & $0.16 \pm 0.01$ & $1.00 \pm 0.00$ \\
Llama-3.1-8B-Instruct & assistant & $33.3 \pm 1.1$ & $0.73 \pm 0.03$ & $0.45 \pm 0.02$ & $0.32 \pm 0.02$ & $0.60 \pm 0.02$ & $1.00 \pm 0.00$ \\
Llama-3.1-8B-Instruct & statement & $26.2 \pm 0.5$ & $0.75 \pm 0.05$ & $0.56 \pm 0.01$ & $0.40 \pm 0.00$ & $0.59 \pm 0.00$ & $1.00 \pm 0.00$ \\
Llama-3.1-8B-Instruct & survey & $18.6 \pm 0.9$ & $0.70 \pm 0.06$ & $0.41 \pm 0.02$ & $0.32 \pm 0.02$ & $0.63 \pm 0.01$ & $1.00 \pm 0.00$ \\
Gemma-2-9B-it & assistant & $59.2 \pm 2.9$ & $0.77 \pm 0.06$ & $0.21 \pm 0.25$ & $0.15 \pm 0.23$ & $0.81 \pm 0.28$ & $1.00 \pm 0.00$ \\
Gemma-2-9B-it & statement & $34.8 \pm 2.2$ & $0.77 \pm 0.07$ & $0.38 \pm 0.01$ & $0.27 \pm 0.01$ & $0.63 \pm 0.03$ & $1.00 \pm 0.00$ \\
Gemma-2-9B-it & survey & $55.9 \pm 1.1$ & $0.80 \pm 0.07$ & $0.38 \pm 0.30$ & $0.26 \pm 0.23$ & $0.67 \pm 0.29$ & $1.00 \pm 0.00$ \\
Gemma-2-27B-it (4-bit) & assistant & $49.0 \pm 2.0$ & $0.80 \pm 0.05$ & $0.20 \pm 0.19$ & $0.12 \pm 0.16$ & $0.89 \pm 0.16$ & $1.00 \pm 0.00$ \\
Gemma-2-27B-it (4-bit) & statement & $47.0 \pm 1.1$ & $0.78 \pm 0.05$ & $0.18 \pm 0.18$ & $0.11 \pm 0.18$ & $0.90 \pm 0.17$ & $1.00 \pm 0.00$ \\
Gemma-2-27B-it (4-bit) & survey & $46.7 \pm 0.5$ & $0.80 \pm 0.05$ & $0.30 \pm 0.14$ & $0.16 \pm 0.19$ & $0.85 \pm 0.19$ & $1.00 \pm 0.00$ \\
Mistral-7B-Instruct-v0.3 & assistant & $45.6 \pm 0.3$ & $0.80 \pm 0.05$ & $0.44 \pm 0.02$ & $0.27 \pm 0.01$ & $0.68 \pm 0.03$ & $1.00 \pm 0.00$ \\
Mistral-7B-Instruct-v0.3 & statement & $33.3 \pm 0.5$ & $0.66 \pm 0.06$ & $0.53 \pm 0.10$ & $0.42 \pm 0.08$ & $0.56 \pm 0.09$ & $1.00 \pm 0.00$ \\
Mistral-7B-Instruct-v0.3 & survey & $48.2 \pm 1.0$ & $0.81 \pm 0.05$ & $0.35 \pm 0.02$ & $0.25 \pm 0.01$ & $0.66 \pm 0.03$ & $1.00 \pm 0.00$ \\
\bottomrule
\end{tabular}
\end{adjustbox}
\end{table}
 
\begin{table}[!htbp]
\centering
\caption{Survey-framing heads under the assistant framing. Head-map $\rho$: Spearman correlation between per-head attribution maps under the two framings. Top-10 share: share of the assistant-framing coupling carried by the survey framing's top-10 heads.}
\label{tab:persona}
\small
\begin{adjustbox}{max width=\linewidth}
\begin{tabular}{lcccc}
\toprule
Model & Pos.\ (\%), survey & Pos.\ (\%), assistant & Head-map $\rho$ & Top-10 share \\
\midrule
Qwen2.5-7B-Instruct & $44.3 \pm 2.4$ & $61.7 \pm 5.5$ & $0.65 \pm 0.06$ & $0.53 \pm 0.21$ \\
Qwen3-4B-Instruct-2507 & $11.7 \pm 1.3$ & $81.7 \pm 3.7$ & $-0.31 \pm 0.05$ & $-0.70 \pm 0.13$ \\
Llama-3.1-8B-Instruct & $18.6 \pm 0.9$ & $30.3 \pm 2.0$ & $0.79 \pm 0.02$ & $0.88 \pm 0.06$ \\
Gemma-2-9B-it & $55.9 \pm 1.1$ & $57.1 \pm 3.7$ & $0.86 \pm 0.02$ & $0.52 \pm 0.16$ \\
Gemma-2-27B-it (4-bit) & $46.7 \pm 0.5$ & $46.8 \pm 2.5$ & $0.83 \pm 0.01$ & $0.52 \pm 0.02$ \\
Mistral-7B-Instruct-v0.3 & $48.2 \pm 1.0$ & $41.9 \pm 0.5$ & $0.81 \pm 0.00$ & $0.61 \pm 0.06$ \\
\bottomrule
\end{tabular}
\end{adjustbox}
\end{table}
 
\FloatBarrier
\section{Ablation Cost and Entanglement}
\label{app:entanglement}
 
All statistics use the band and the band direction $g$ of the control stage. Enrichment is the variance of ordinary text along $g$ divided by the average variance per direction (the trace of the covariance divided by the hidden size), computed over all tokens of 100 WikiText-2 passages except the first; the same quantity for a random unit direction is close to 1 in every model. Sink shift is the mean squared distance between the first token's projection on $g$ and the mean projection $\mu$, relative to the variance of ordinary tokens along $g$. Principal components are computed from 4{,}000 ordinary-text token states per layer. All values are band means. Perplexity and MMLU change in Table~\ref{tab:e1} are from the control stage; Table~\ref{tab:e3} uses the 200-pair sink-test sample.
 
\begin{table}[!htbp]
\centering
\caption{Predictors of ablation cost (means $\pm$ s.d.\ over seeds). Base models have no control stage.}
\label{tab:e1}
\scriptsize
\setlength{\tabcolsep}{3pt}
\begin{adjustbox}{max width=\linewidth}
\begin{tabular}{lcccccccc}
\toprule
Model & Enrichment & Random dir. & Max enrichment & Sink shift & First-token norm ratio & $\cos^2$ PC1 & PPL ratio & MMLU change \\
\midrule
Qwen2.5-0.5B-Instruct & $2.2 \pm 0.0$ & $1.0 \pm 0.0$ & $2.4 \pm 0.0$ & $3252.2 \pm 7.5$ & $88.1 \pm 0.0$ & $0.001 \pm 0.000$ & $1.028 \pm 0.002$ & $-0.001 \pm 0.001$ \\
Qwen2.5-1.5B-Instruct & $2.7 \pm 0.0$ & $1.0 \pm 0.0$ & $4.3 \pm 0.1$ & $14912.0 \pm 591.4$ & $137.3 \pm 0.0$ & $0.002 \pm 0.000$ & $1.023 \pm 0.000$ & $-0.022 \pm 0.003$ \\
Qwen2.5-3B-Instruct & $2.3 \pm 0.3$ & $1.0 \pm 0.3$ & $4.9 \pm 1.6$ & $639.0 \pm 231.4$ & $34.6 \pm 2.5$ & $0.001 \pm 0.000$ & $1.011 \pm 0.006$ & $0.003 \pm 0.003$ \\
Qwen2.5-7B-Instruct & $10.6 \pm 0.5$ & $1.6 \pm 0.6$ & $62.7 \pm 3.4$ & $500.0 \pm 7.6$ & $80.2 \pm 0.0$ & $0.003 \pm 0.000$ & $1.040 \pm 0.001$ & $0.001 \pm 0.003$ \\
Qwen2.5-7B$^\dagger$ & $6.4 \pm 0.1$ & $1.8 \pm 0.7$ & $16.5 \pm 0.4$ & $507.2 \pm 7.5$ & $88.8 \pm 0.0$ & $0.002 \pm 0.000$ & -- & -- \\
Qwen2.5-14B-Instruct & $4.7 \pm 0.2$ & $1.0 \pm 0.0$ & $6.2 \pm 0.3$ & $92488.8 \pm 8088.7$ & $145.3 \pm 7.3$ & $0.004 \pm 0.001$ & $1.074 \pm 0.017$ & $-0.001 \pm 0.004$ \\
Qwen2.5-32B-Instruct & $3.5 \pm 0.0$ & $1.0 \pm 0.0$ & $4.3 \pm 0.1$ & $11859.6 \pm 2602.1$ & $98.9 \pm 0.0$ & $0.001 \pm 0.000$ & $1.026 \pm 0.005$ & $-0.001 \pm 0.001$ \\
Qwen2.5-72B-Instruct (4-bit) & $2.6 \pm 0.1$ & $1.0 \pm 0.0$ & $3.8 \pm 0.0$ & $2059.3 \pm 194.8$ & $26.2 \pm 2.3$ & $0.001 \pm 0.000$ & $1.011 \pm 0.004$ & $-0.004 \pm 0.000$ \\
Qwen3-0.6B & $2.0 \pm 0.0$ & $1.0 \pm 0.0$ & $3.3 \pm 0.0$ & $2189.4 \pm 70.2$ & $48.7 \pm 0.0$ & $0.002 \pm 0.000$ & $1.148 \pm 0.009$ & $-0.005 \pm 0.001$ \\
Qwen3-1.7B & $4.2 \pm 0.1$ & $1.0 \pm 0.0$ & $8.3 \pm 0.4$ & $1828.5 \pm 41.8$ & $42.6 \pm 0.0$ & $0.014 \pm 0.001$ & $1.844 \pm 0.081$ & $-0.009 \pm 0.001$ \\
Qwen3-4B & $7.1 \pm 0.2$ & $1.1 \pm 0.3$ & $22.8 \pm 0.9$ & $3772.2 \pm 114.4$ & $97.8 \pm 0.0$ & $0.004 \pm 0.000$ & $1.395 \pm 0.014$ & $-0.018 \pm 0.002$ \\
Qwen3-4B-Base$^\dagger$ & $1.5 \pm 0.0$ & $1.0 \pm 0.2$ & $1.9 \pm 0.0$ & $599.7 \pm 111.6$ & $18.3 \pm 0.0$ & $0.000 \pm 0.000$ & -- & -- \\
Qwen3-4B-Instruct-2507 & $3.0 \pm 0.1$ & $1.1 \pm 0.2$ & $4.1 \pm 0.2$ & $2792.5 \pm 265.6$ & $77.0 \pm 0.0$ & $0.001 \pm 0.000$ & $1.006 \pm 0.005$ & $0.001 \pm 0.004$ \\
Qwen3-8B & $4.5 \pm 0.3$ & $0.8 \pm 0.0$ & $13.5 \pm 0.2$ & $8702.5 \pm 4194.5$ & $85.5 \pm 18.0$ & $0.001 \pm 0.000$ & $1.024 \pm 0.009$ & $-0.011 \pm 0.006$ \\
Qwen3-14B & $6.8 \pm 0.1$ & $1.0 \pm 0.1$ & $29.9 \pm 0.7$ & $20015.5 \pm 138.1$ & $91.4 \pm 0.0$ & $0.002 \pm 0.000$ & $1.050 \pm 0.001$ & $-0.009 \pm 0.001$ \\
Qwen3-32B & $22.8 \pm 9.6$ & $0.9 \pm 0.1$ & $84.9 \pm 24.4$ & $3586.5 \pm 342.2$ & $76.0 \pm 6.0$ & $0.008 \pm 0.004$ & $1.785 \pm 0.374$ & $0.003 \pm 0.002$ \\
Llama-3.1-8B-Instruct & $2.7 \pm 0.0$ & $1.0 \pm 0.0$ & $4.0 \pm 0.1$ & $6571.4 \pm 90.3$ & $77.1 \pm 0.0$ & $0.002 \pm 0.000$ & $1.026 \pm 0.001$ & $-0.007 \pm 0.002$ \\
Llama-3.1-70B-Instruct (4-bit) & $2.6 \pm 0.2$ & $1.0 \pm 0.0$ & $9.9 \pm 0.4$ & $3938.0 \pm 380.7$ & $46.9 \pm 4.6$ & $0.001 \pm 0.000$ & $1.046 \pm 0.007$ & $0.007 \pm 0.001$ \\
Gemma-2-9B-it & $4.0 \pm 0.2$ & $1.0 \pm 0.0$ & $33.2 \pm 1.4$ & $625.6 \pm 42.7$ & $15.6 \pm 0.3$ & $0.003 \pm 0.001$ & $1.001 \pm 0.006$ & $0.003 \pm 0.003$ \\
Gemma-2-27B-it & $62.9 \pm 0.4$ & $0.9 \pm 0.1$ & $247.2 \pm 2.7$ & $750.3 \pm 6.5$ & $10.1 \pm 0.3$ & $0.024 \pm 0.001$ & $13.927 \pm 2.532$ & $-0.244 \pm 0.004$ \\
Mistral-7B-Instruct-v0.3 & $1.8 \pm 0.1$ & $1.0 \pm 0.0$ & $2.5 \pm 0.0$ & $4895.0 \pm 1191.3$ & $51.2 \pm 9.1$ & $0.000 \pm 0.000$ & $0.998 \pm 0.001$ & $0.001 \pm 0.001$ \\
Mistral-7B-v0.3$^\dagger$ & $2.1 \pm 0.0$ & $1.0 \pm 0.0$ & $2.9 \pm 0.0$ & $8435.9 \pm 3884.7$ & $41.6 \pm 19.0$ & $0.001 \pm 0.000$ & -- & -- \\
\bottomrule
\end{tabular}
\end{adjustbox}
\end{table}
 
\begin{table}[!htbp]
\centering
\caption{Spearman correlation across the 19 instruction-tuned models between each predictor and the control-stage cost (4-bit checks excluded); permutation $p$-values from 10{,}000 permutations.}
\label{tab:e2}
\small
\begin{adjustbox}{max width=\linewidth}
\begin{tabular}{lcccc}
\toprule
Predictor & $\rho$ (log PPL ratio) & $p$ & $\rho$ (MMLU change) & $p$ \\
\midrule
Enrichment & $0.56$ & $p    0.0131
p    0.2855
Name: 0, dtype: str$ & $-0.26$ & $0.2855$ \\
Max enrichment in band & $0.45$ & $p    0.0532
p    0.9036
Name: 1, dtype: str$ & $-0.03$ & $0.9036$ \\
Sink shift & $0.01$ & $p    0.9671
p    0.2383
Name: 2, dtype: str$ & $-0.28$ & $0.2383$ \\
First-position norm / typical & $0.06$ & $p    0.8063
p    0.2588
Name: 3, dtype: str$ & $-0.27$ & $0.2588$ \\
$\cos^2$ with PC1 & $0.73$ & $p    0.0006
p    0.1939
Name: 4, dtype: str$ & $-0.31$ & $0.1939$ \\
$\cos^2$ with top-10 PCs & $0.60$ & $p    0.0086
p    0.0971
Name: 5, dtype: str$ & $-0.39$ & $0.0971$ \\
\bottomrule
\end{tabular}
\end{adjustbox}
\end{table}
 
\begin{table}[!htbp]
\centering
\caption{Band ablation of $g$ at every position, with the first token spared, and with every position whose norm exceeds ten times the median spared (200 held-out pairs, 100 passages, 200 MMLU questions per seed; means over seeds). $\kappa_g$: share of coupling remaining. PPL: perplexity ratio. MMLU: change in accuracy (points). Ans.: standalone answers that keep their sign (\%). Massive: positions above the norm threshold (\%).}
\label{tab:e3}
\scriptsize
\setlength{\tabcolsep}{2.5pt}
\begin{adjustbox}{max width=\linewidth}
\begin{tabular}{lccccccccccccc}
\toprule
 & \multicolumn{4}{c}{All positions} & \multicolumn{4}{c}{First token spared} & \multicolumn{4}{c}{Massive positions spared} & \\
\cmidrule(lr){2-5}\cmidrule(lr){6-9}\cmidrule(lr){10-13}
Model & $\kappa_g$ & PPL & MMLU & Ans. & $\kappa_g$ & PPL & MMLU & Ans. & $\kappa_g$ & PPL & MMLU & Ans. & Massive \\
\midrule
Qwen2.5-0.5B-Instruct & $0.00$ & $1.026$ & $-0.7$ & $59$ & $0.00$ & $1.005$ & $-0.3$ & $78$ & $0.00$ & $1.005$ & $-0.3$ & $78$ & $1.32$ \\
Qwen2.5-1.5B-Instruct & $0.07$ & $1.023$ & $-3.7$ & $96$ & $0.08$ & $1.006$ & $-3.5$ & $99$ & $0.08$ & $1.006$ & $-3.3$ & $99$ & $1.24$ \\
Qwen2.5-3B-Instruct & $0.10$ & $1.011$ & $-1.0$ & $93$ & $0.11$ & $1.001$ & $-0.7$ & $94$ & $0.11$ & $1.000$ & $-0.5$ & $94$ & $1.13$ \\
Qwen2.5-7B-Instruct & $0.13$ & $1.041$ & $+1.0$ & $95$ & $0.13$ & $1.009$ & $+1.2$ & $95$ & $0.14$ & $0.999$ & $+1.3$ & $96$ & $1.33$ \\
Qwen2.5-7B$^\dagger$ & $0.15$ & $1.018$ & $+0.0$ & $84$ & $0.15$ & $1.007$ & $+0.3$ & $84$ & $0.15$ & $1.003$ & $+0.0$ & $87$ & $1.33$ \\
Qwen2.5-14B-Instruct & $0.06$ & $1.084$ & $+1.0$ & $96$ & $0.06$ & $1.001$ & $+0.8$ & $96$ & $0.07$ & $1.001$ & $+0.2$ & $99$ & $1.32$ \\
Qwen2.5-32B-Instruct & $0.14$ & $1.028$ & $-0.2$ & $96$ & $0.13$ & $1.002$ & $+0.0$ & $96$ & $0.13$ & $1.002$ & $+0.5$ & $97$ & $1.32$ \\
Qwen2.5-72B-Instruct (4-bit) & $0.10$ & $1.011$ & $+0.0$ & $92$ & $0.12$ & $1.001$ & $+0.3$ & $98$ & $0.12$ & $1.002$ & $+0.3$ & $98$ & $1.26$ \\
Qwen3-0.6B & $0.06$ & $1.146$ & $-0.8$ & $30$ & $0.07$ & $1.003$ & $+1.3$ & $91$ & $0.07$ & $1.007$ & $+1.3$ & $91$ & $1.26$ \\
Qwen3-1.7B & $0.07$ & $1.858$ & $-1.3$ & $72$ & $0.13$ & $1.003$ & $+0.7$ & $60$ & $0.12$ & $1.210$ & $+0.5$ & $83$ & $1.16$ \\
Qwen3-4B & $0.03$ & $1.438$ & $-1.5$ & $66$ & $0.05$ & $1.013$ & $+1.3$ & $66$ & $0.05$ & $1.000$ & $+1.3$ & $66$ & $1.28$ \\
Qwen3-4B-Base$^\dagger$ & $-0.15$ & $1.013$ & $+1.7$ & $98$ & $-0.15$ & $1.001$ & $+0.5$ & $100$ & $-0.15$ & $1.007$ & $+0.0$ & $100$ & $1.09$ \\
Qwen3-4B-Instruct-2507 & $0.02$ & $1.010$ & $+0.2$ & $94$ & $0.02$ & $1.002$ & $+0.5$ & $92$ & $0.02$ & $1.002$ & $+0.5$ & $92$ & $1.33$ \\
Qwen3-8B & $0.12$ & $1.024$ & $-3.3$ & $96$ & $0.12$ & $1.001$ & $-1.5$ & $94$ & $0.12$ & $1.001$ & $-1.5$ & $94$ & $1.27$ \\
Qwen3-14B & $0.09$ & $1.049$ & $-1.5$ & $95$ & $0.10$ & $1.004$ & $-0.8$ & $96$ & $0.10$ & $1.004$ & $-0.8$ & $96$ & $2.04$ \\
Qwen3-32B & $0.07$ & $1.806$ & $-0.3$ & $93$ & $0.14$ & $1.002$ & $-0.3$ & $95$ & $0.14$ & $0.997$ & $-0.3$ & $95$ & $1.27$ \\
Llama-3.1-8B-Instruct & $0.02$ & $1.028$ & $+0.3$ & $94$ & $0.04$ & $1.002$ & $-0.8$ & $94$ & $0.04$ & $1.002$ & $-0.8$ & $94$ & $1.05$ \\
Llama-3.1-70B-Instruct (4-bit) & $0.10$ & $1.055$ & $+0.2$ & $94$ & $0.11$ & $1.003$ & $-0.2$ & $91$ & $0.11$ & $1.003$ & $-0.2$ & $91$ & $1.05$ \\
Gemma-2-9B-it & $0.03$ & $1.005$ & $+0.8$ & $98$ & $0.04$ & $1.002$ & $+1.0$ & $98$ & $0.04$ & $1.002$ & $+1.0$ & $98$ & $1.35$ \\
Gemma-2-27B-it & $0.00$ & $15.401$ & $-23.2$ & $83$ & $0.00$ & $1.101$ & $-1.7$ & $94$ & $0.00$ & $2.606$ & $-10.5$ & $91$ & $0.36$ \\
Mistral-7B-Instruct-v0.3 & $-0.02$ & $0.997$ & $-1.3$ & $94$ & $-0.03$ & $1.000$ & $-1.0$ & $94$ & $-0.03$ & $1.001$ & $-1.2$ & $94$ & $2.43$ \\
Mistral-7B-v0.3$^\dagger$ & $0.10$ & $1.014$ & $-1.2$ & $88$ & $0.11$ & $1.000$ & $-0.3$ & $76$ & $0.11$ & $1.000$ & $-1.2$ & $74$ & $2.46$ \\
\bottomrule
\end{tabular}
\end{adjustbox}
\end{table}
 
\FloatBarrier
\section{Trade-off Sweep}
\label{app:tradeoff}
 
The sweep varies the readout band's profile thresholds (0.9--0.1, the default; 0.75--0.25; 0.6--0.4, narrowest) and the ablation strength $\alpha$, with ablation at every position (200 held-out pairs, 200 MMLU questions, 100 passages per seed). When two thresholds give the same band in a seed, the setting is evaluated once; a setting without a standard deviation was available in one seed only. The chosen setting is selected by leave-one-seed-out: for each seed, the setting with the lowest mean perplexity ratio over the other two seeds among those whose mean coupling remaining is at most 0.2, reported on the held-out seed.
 
\begin{table}[!htbp]
\centering
\caption{Every setting: share of coupling remaining / perplexity increase (\%), means over seeds. Dashes: setting not evaluated.}
\label{tab:tradeoff_full}
\scriptsize
\setlength{\tabcolsep}{2pt}
\begin{adjustbox}{max width=\linewidth}
\begin{tabular}{lccccccccc}
\toprule
 & \multicolumn{3}{c}{Band 0.9--0.1} & \multicolumn{3}{c}{Band 0.75--0.25} & \multicolumn{3}{c}{Band 0.6--0.4} \\
\cmidrule(lr){2-4}\cmidrule(lr){5-7}\cmidrule(lr){8-10}
Model & $\alpha=0.5$ & $0.75$ & $1$ & $0.5$ & $0.75$ & $1$ & $0.5$ & $0.75$ & $1$ \\
\midrule
Qwen2.5-0.5B-Instruct & 0.14 / 2.0 & 0.03 / 2.8 & 0.00 / 2.6 & 0.33 / 0.6 & 0.20 / 1.6 & 0.16 / 1.9 & 0.36 / 1.5 & 0.21 / 1.2 & 0.16 / 0.4 \\
Qwen2.5-1.5B-Instruct & 0.11 / 2.5 & 0.09 / 2.4 & 0.07 / 2.3 & 0.35 / 1.4 & 0.24 / 1.7 & 0.15 / 1.8 & 0.42 / 0.7 & 0.24 / 1.6 & 0.14 / 2.2 \\
Qwen2.5-3B-Instruct & 0.20 / 0.6 & 0.13 / 0.9 & 0.10 / 1.1 & 0.45 / 0.5 & 0.35 / 1.2 & 0.27 / 2.1 & 0.49 / 0.7 & 0.37 / 1.4 & 0.28 / 2.4 \\
Qwen2.5-7B-Instruct & 0.16 / 2.3 & 0.14 / 3.5 & 0.13 / 4.1 & 0.38 / 2.2 & 0.32 / 3.0 & 0.29 / 3.9 & 0.57 / 1.5 & 0.45 / 2.1 & 0.35 / 3.4 \\
Qwen2.5-14B-Instruct & 0.08 / 6.2 & 0.06 / 7.5 & 0.06 / 8.4 & 0.18 / 6.8 & 0.16 / 7.9 & 0.15 / 8.6 & 0.28 / 6.2 & 0.26 / 7.0 & 0.25 / 7.3 \\
Qwen2.5-32B-Instruct & 0.16 / 1.6 & 0.14 / 2.2 & 0.14 / 2.8 & 0.40 / 1.2 & 0.36 / 1.8 & 0.35 / 2.2 & 0.36 / 0.6 & 0.32 / 1.0 & 0.31 / 1.5 \\
Qwen2.5-32B-Instruct (4-bit) & 0.18 / 0.9 & 0.15 / 1.5 & 0.14 / 1.9 & -- & -- & -- & 0.41 / 0.4 & 0.36 / 0.6 & 0.35 / 0.9 \\
Qwen3-0.6B & 0.12 / 9.2 & 0.07 / 12.5 & 0.06 / 14.6 & 0.12 / 7.0 & 0.07 / 12.5 & 0.06 / 16.7 & -- & -- & -- \\
Qwen3-1.7B & 0.13 / 53.0 & 0.09 / 73.3 & 0.07 / 85.8 & 0.24 / 11.4 & 0.18 / 19.4 & 0.15 / 26.6 & 0.28 / -0.6 & 0.22 / -0.9 & 0.19 / -1.0 \\
Qwen3-4B & 0.18 / 30.0 & 0.11 / 37.9 & 0.03 / 43.8 & 0.19 / 15.2 & 0.12 / 21.8 & 0.08 / 26.2 & 0.22 / 16.4 & 0.18 / 23.2 & 0.15 / 27.2 \\
Qwen3-4B-Instruct-2507 & 0.06 / 0.9 & 0.04 / 0.9 & 0.02 / 1.0 & 0.21 / 0.8 & 0.15 / 0.8 & 0.13 / 0.6 & 0.28 / 1.0 & 0.21 / 1.1 & 0.19 / 1.0 \\
Qwen3-8B & 0.24 / 0.7 & 0.17 / 1.8 & 0.12 / 2.4 & 0.31 / 0.8 & 0.22 / 1.8 & 0.15 / 2.5 & 0.29 / -0.3 & 0.20 / -0.1 & 0.15 / -0.1 \\
Qwen3-14B & 0.28 / 1.8 & 0.16 / 3.2 & 0.09 / 4.9 & 0.21 / 1.1 & 0.12 / 1.5 & 0.07 / 2.2 & 0.28 / 0.9 & 0.19 / 0.6 & 0.16 / 0.7 \\
Qwen3-32B & 0.26 / 49.1 & 0.15 / 68.2 & 0.07 / 80.6 & 0.39 / 18.7 & 0.26 / 28.5 & 0.18 / 37.0 & -- & -- & -- \\
Qwen3-32B (4-bit) & 0.25 / 48.2 & 0.13 / 67.8 & 0.07 / 80.6 & 0.34 / 29.6 & 0.21 / 44.6 & 0.14 / 56.8 & 0.68 / 12.4 & 0.61 / 21.4 & 0.54 / 32.1 \\
Llama-3.1-8B-Instruct & 0.08 / 2.1 & 0.04 / 2.6 & 0.02 / 2.8 & 0.10 / 1.2 & 0.05 / 1.5 & 0.04 / 1.6 & 0.20 / 0.9 & 0.13 / 1.1 & 0.10 / 1.2 \\
Gemma-2-9B-it & 0.07 / 0.7 & 0.04 / 0.6 & 0.03 / 0.5 & 0.20 / 0.6 & 0.18 / 0.6 & 0.17 / 0.6 & 0.34 / 0.3 & 0.32 / 0.2 & 0.32 / 0.1 \\
Gemma-2-27B-it & 0.00 / 616.7 & 0.00 / 1048.3 & 0.00 / 1440.1 & 0.07 / 439.5 & 0.04 / 714.2 & 0.02 / 995.5 & 0.32 / 39.4 & 0.26 / 63.4 & 0.22 / 89.0 \\
Gemma-2-27B-it (4-bit) & 0.01 / 714.4 & 0.00 / 1120.8 & 0.00 / 1527.0 & 0.08 / 444.9 & 0.06 / 661.5 & 0.04 / 897.5 & 0.30 / 293.8 & 0.27 / 423.8 & 0.25 / 518.6 \\
Mistral-7B-Instruct-v0.3 & 0.05 / -0.2 & 0.01 / -0.2 & 0.02 / -0.3 & 0.20 / 0.0 & 0.18 / 0.0 & 0.17 / 0.0 & 0.52 / 0.0 & 0.42 / 0.0 & 0.37 / -0.1 \\
\bottomrule
\end{tabular}
\end{adjustbox}
\end{table}
 
\begin{table}[!htbp]
\centering
\caption{Chosen setting per model (leave-one-seed-out; several settings when seeds chose differently) and its effect on the held-out seed. Ans.: standalone answers that keep their sign (\%).}
\label{tab:tradeoff_chosen}
\scriptsize
\begin{adjustbox}{max width=\linewidth}
\begin{tabular}{llccccc}
\toprule
Model & Chosen settings & $\kappa_g$ & PPL ratio & MMLU change & Ans. & Random dir. \\
\midrule
Qwen2.5-0.5B-Instruct & 0.6-0.4 $\alpha$ 1 & $0.16 \pm 0.03$ & $1.004 \pm 0.000$ & $-0.025 \pm 0.007$ & $93.2 \pm 1.1$ & $1.02 \pm 0.00$ \\
Qwen2.5-1.5B-Instruct & 0.75-0.25 $\alpha$ 1 & $0.15 \pm 0.03$ & $1.018 \pm 0.001$ & $-0.017 \pm 0.008$ & $99.4 \pm 1.0$ & $0.95 \pm 0.04$ \\
Qwen2.5-3B-Instruct & 0.9-0.1 $\alpha$ 0.5; 0.9-0.1 $\alpha$ 0.75 & $0.16 \pm 0.10$ & $1.008 \pm 0.006$ & $-0.012 \pm 0.006$ & $93.8 \pm 1.8$ & $1.01 \pm 0.05$ \\
Qwen2.5-7B-Instruct & 0.9-0.1 $\alpha$ 0.5 & $0.16 \pm 0.01$ & $1.023 \pm 0.001$ & $0.008 \pm 0.003$ & $91.0 \pm 2.1$ & $0.99 \pm 0.03$ \\
Qwen2.5-14B-Instruct & 0.9-0.1 $\alpha$ 0.5 & $0.08 \pm 0.03$ & $1.062 \pm 0.015$ & $0.010 \pm 0.000$ & $97.2 \pm 1.7$ & $1.01 \pm 0.04$ \\
Qwen2.5-32B-Instruct & 0.9-0.1 $\alpha$ 0.5 & $0.16 \pm 0.04$ & $1.016 \pm 0.004$ & $0.000 \pm 0.000$ & $96.1 \pm 1.0$ & $1.00 \pm 0.01$ \\
Qwen2.5-32B-Instruct (4-bit) & 0.9-0.1 $\alpha$ 0.5 & $0.18 \pm 0.02$ & $1.009 \pm 0.005$ & $-0.005 \pm 0.000$ & $97.2 \pm 1.8$ & $1.00 \pm 0.01$ \\
Qwen3-0.6B & 0.75-0.25 $\alpha$ 0.5 & $0.12 \pm 0.00$ & $1.070 \pm 0.004$ & $-0.008 \pm 0.003$ & $30.7 \pm 1.5$ & $1.19 \pm 0.06$ \\
Qwen3-1.7B & 0.6-0.4 $\alpha$ 1; 0.75-0.25 $\alpha$ 0.75 & $0.17$ & $1.605$ & $-0.010$ & $71.4$ & $0.97 \pm 0.02$ \\
Qwen3-4B & 0.75-0.25 $\alpha$ 0.5; 0.75-0.25 $\alpha$ 0.75 & $0.17 \pm 0.05$ & $1.173 \pm 0.032$ & $-0.017 \pm 0.008$ & $65.0 \pm 9.7$ & $0.98 \pm 0.11$ \\
Qwen3-4B-Instruct-2507 & 0.75-0.25 $\alpha$ 1 & $0.13 \pm 0.06$ & $1.006 \pm 0.005$ & $0.008 \pm 0.003$ & $92.9 \pm 1.4$ & $0.94 \pm 0.04$ \\
Qwen3-8B & 0.6-0.4 $\alpha$ 1 & $0.15 \pm 0.02$ & $0.999 \pm 0.000$ & $-0.010 \pm 0.007$ & $96.6 \pm 1.2$ & $1.01 \pm 0.01$ \\
Qwen3-14B & 0.6-0.4 $\alpha$ 0.75; 0.6-0.4 $\alpha$ 1 & $0.17 \pm 0.01$ & $1.011 \pm 0.013$ & $0.000 \pm 0.005$ & $87.6 \pm 2.6$ & $0.93 \pm 0.04$ \\
Qwen3-32B & 0.75-0.25 $\alpha$ 1; 0.9-0.1 $\alpha$ 0.75 & $0.17 \pm 0.06$ & $1.550 \pm 0.369$ & $-0.003 \pm 0.008$ & $94.9 \pm 0.9$ & $1.01 \pm 0.05$ \\
Qwen3-32B (4-bit) & 0.75-0.25 $\alpha$ 0.75; 0.75-0.25 $\alpha$ 1; 0.9-0.1 $\alpha$ 0.75 & $0.15 \pm 0.08$ & $1.711 \pm 0.370$ & $0.002 \pm 0.010$ & $94.6 \pm 1.8$ & $1.01 \pm 0.03$ \\
Llama-3.1-8B-Instruct & 0.6-0.4 $\alpha$ 0.5; 0.6-0.4 $\alpha$ 0.75 & $0.16 \pm 0.17$ & $1.010 \pm 0.002$ & $0.002 \pm 0.006$ & $96.6 \pm 2.3$ & $0.97 \pm 0.05$ \\
Gemma-2-9B-it & 0.75-0.25 $\alpha$ 1; 0.9-0.1 $\alpha$ 1 & $0.16 \pm 0.12$ & $1.008 \pm 0.004$ & $0.005 \pm 0.000$ & $97.2 \pm 2.0$ & $1.00 \pm 0.01$ \\
Gemma-2-27B-it & 0.6-0.4 $\alpha$ 1; 0.75-0.25 $\alpha$ 0.5 & $0.19 \pm 0.16$ & $3.946 \pm 3.044$ & $-0.042 \pm 0.046$ & $92.7 \pm 3.9$ & $0.93 \pm 0.05$ \\
Gemma-2-27B-it (4-bit) & 0.75-0.25 $\alpha$ 0.5 & $0.08 \pm 0.07$ & $5.449 \pm 1.661$ & $-0.080 \pm 0.035$ & $94.4 \pm 5.4$ & $0.89 \pm 0.03$ \\
Mistral-7B-Instruct-v0.3 & 0.9-0.1 $\alpha$ 1 & $-0.02 \pm 0.01$ & $0.997 \pm 0.001$ & $-0.013 \pm 0.003$ & $94.4 \pm 4.2$ & $1.00 \pm 0.02$ \\
\bottomrule
\end{tabular}
\end{adjustbox}
\end{table}
 
\FloatBarrier
\section{Implementation}
\label{app:suite}
 
The experiments are implemented in a single resumable suite, provided in the supplementary material. Each run is described by a JSON configuration listing models, seeds, framings and stages; every (model, seed, framing, stage) result is stored as a separate file and skipped if present, and a run refuses to mix results computed with different parameters. The stages are \texttt{behaviour} (coupling and geometry), \texttt{causal} (patches and ablations at $l^*$, layer profile, exactness checks), \texttt{readout} (band removal and read gain), \texttt{heads} (attribution and validation), \texttt{control} (interventions at every position), \texttt{persona} (heads under the assistant framing), \texttt{tradeoff} (band and strength sweep) and \texttt{pairs} (per-pair records and confidence bins). A separate script computes the entanglement statistics and the sink test from the stored results, and a report script aggregates all results into the tables and figures. Models with a heads stage use eager attention. Random-weight controls re-initialise the same architecture with a fixed seed and run the behaviour stage only. Experiments ran on two NVIDIA RTX 6000 Ada GPUs (48\,GB each); bfloat16 models of 27B and above were split across both GPUs, and 4-bit models used bitsandbytes NF4 with double quantisation and bfloat16 computation on one GPU.
 
\FloatBarrier

\end{document}